\documentclass[10pt,a4paper]{article}

\usepackage{arxiv}
\usepackage{tikz}
\usepackage[utf8]{inputenc} 
\usepackage[T1]{fontenc}    
\usepackage{hyperref}       
\usepackage{url}            
\usepackage{booktabs}       
\usepackage{amsfonts}       
\usepackage{nicefrac}       
\usepackage{microtype}      
\usepackage{lipsum}

\usepackage{graphicx}
\usepackage{multirow}
\usepackage[misc]{ifsym}
\usepackage{subfigure}
\usepackage{natbib}
\usepackage{color}
\usepackage{amsmath}
\usepackage{bm}
\usepackage[noend]{algpseudocode}
\usepackage{algorithmicx,algorithm}
\usetikzlibrary{positioning}

\DeclareFontFamily{OT1}{pzc}{}
\DeclareFontShape{OT1}{pzc}{m}{it}{<-> s * [1.10] pzcmi7t}{}
\DeclareMathAlphabet{\mathpzc}{OT1}{pzc}{m}{it}
\usepackage{natbib}
\title{Physics-informed Convolutional Neural Networks for Temperature Field Prediction of Heat Source Layout without Labeled Data}

\author{
  Xiaoyu Zhao \\
  Defense Innovation Institute \\
  Chinese Academy of Military Science, 
  Beijing, China \\
  \texttt{zhaoxiaoyu13@nudt.edu.cn} \\
  \AND
  Zhiqiang Gong \\
  Defense Innovation Institute \\
  Chinese Academy of Military Science, 
  Beijing, China \\
  \texttt{gongzhiqiang13@nudt.edu.cn} \\
  \AND
  Yunyang Zhang \\
  Defense Innovation Institute \\
  Chinese Academy of Military Science, 
  Beijing, China \\
  \texttt{zhangyunyang17@csu.ac.cn} \\
  \AND
  Wen Yao \\
  Defense Innovation Institute \\
  Chinese Academy of Military Science, 
  Beijing, China \\
  \texttt{wendy0782@126.com} \\
  \AND
  Xiaoqian Chen \\
  Defense Innovation Institute \\
  Chinese Academy of Military Science, 
  Beijing, China \\
  \texttt{chenxiaoqian@nudt.edu.cn} \\
  }

\begin{document}
\maketitle

\begin{abstract}
Recently, surrogate models based on deep learning have attracted much attention for engineering analysis and optimization.
As the construction of data pairs in most engineering problems is time-consuming, data acquisition is becoming the predictive capability bottleneck of most deep surrogate models, which also exists in surrogate for thermal analysis and design.
To address this issue, this paper develops a physics-informed convolutional neural network (CNN) for the thermal simulation surrogate.
The network can learn a mapping from heat source layout to the steady-state temperature field without labeled data, which equals solving an entire family of partial difference equations (PDEs).
To realize the physics-guided training without labeled data, we employ the heat conduction equation and finite difference method to construct the loss function. 
Since the solution is sensitive to boundary conditions, we properly impose hard constraints by padding in the Dirichlet and Neumann boundary conditions.
In addition, the neural network architecture is well-designed to improve the prediction precision of the problem at hand, and pixel-level online hard example mining is introduced to overcome the imbalance of optimization difficulty in the computation domain.
The experiments demonstrate that the proposed method can provide comparable predictions with numerical method and data-driven deep learning models. We also conduct various ablation studies to investigate the effectiveness of the network component and training methods proposed in this paper.  

\end{abstract}

\keywords{Physics-informed convolutional neural networks \and Thermal simulation \and Heat source layout \and Finite difference method \and Temperature field prediction}

\section{Introduction}
\label{intro}
With the increasing integration of electronic equipment, thermal management becomes an essential part of the electronic system design to ensure the performance and reliability of electronic components. 
One effective way for thermal management is to optimize the heat source layout inside the electronic device (\cite{chen2016};\cite{chen2017};\cite{chen2018})), where physics simulation plays an important role to provide temperature field computations for layout evaluation in the layout optimization loop.
The physics simulation usually involves solving complex partial differential equations (PDEs).
It is expensive and time-consuming in many science and engineering problems, including the thermal analysis and design, as PDEs with different design parameters must be repeatedly solved.
Surrogate models, such as Gaussian processes (\cite{lee2020propagation}), kriging (\cite{liu2017adaptive}), SVM regression (\cite{moustapha2018comparative}), and neural network (\cite{misiulia2017geometry}), are computationally attractive to replace the costly and repetitive physics simulations. Recently, more researches focused on the deep learning-based surrogate model benefit from the power of deep neural networks (DNNs) in solving high-dimensional and strong nonlinear problems.

This paper studies the deep surrogate model in one thermal management task named temperature field prediction of heat source layout (HSL-TFP).
There exist some works on the data-driven surrogate model based on DNNs. \cite{chen2020heat} modeled the temperature field prediction as an image-to-image regression task, and feature pyramid network is employed as the backbone of DNNs. Compared with numerical methods, the DNNs surrogate reduces the computation time and improves the efficiency of layout optimization.
For evaluating the performance of various methods, \cite{chen2021deep} summarized the HSL-TFP task benchmark from data generation, baseline models, and evaluation metrics.
The data-driven deep surrogate model are also investigated in structural mechanics (\cite{kallioras2020accelerated};\cite{ates2021two}), fluid mechanics (\cite{zhu2019physics};\cite{li2020fourier}), and materials physics (\cite{teichert2019machine}).
As deep learning methods operate in the big data regime, amounts of training data are usually necessary to achieve reliable predictive performance.
However, preparing sufficient simulation data is similarly expensive and time-consuming in such engineering and scientific field.

To reduce the cost of preparing simulation data, we investigate to incorporate physical knowledge into DNN training. 
Some similar research exists called physics-informed neural network (PINN), or physics-constrained neural network.
Classical PINN works primarily focus on solving one PDE with specific parameters by fully-connected neural networks (FC-NNs). 
For example, \cite{raissi2019physics} exploited PINN with the development of deep learning techniques. It leverages the capability of the deep neural network as universal function approximators and the techniques of automatic differentiation (\cite{baydin2018automatic}). The physics-informed loss function of PINN is constructed based on governing PDEs and is used to regularize the output for minimizing the violation of PDEs.
This representive PINN has been studied in many scientific problems to solve various PDEs, including high-speed flows (\cite{mao2020physics}), subsurface flows (\cite{wang2020deep}), metamaterial design (\cite{liu2019multi}), power system (\cite{misyris2020physics}), and heat transfer equation (\cite{zobeiry2021physics}).

In contrast with these studies, our work focuses on physics-informed surrogate model to learn the mapping from parametric dependence to PDE solution. It will learn an entire family of PDEs, which is more challenging than solving one instance of PDE. 
Therefore, we present the physics-informed surrogate model for HSL-TFP with CNN architecture. The input and output are the discrete representation of parameter and solution function, which can be treated as images.
Different from automatic differentiation, the finite difference method is utilized to join the heat conduction equation into the construction of loss function. Besides, the hard constraint is imposed in the boundary by padding to avoid ill-posed solution. They collectively guide the network to minimize the violation of equations. 
In fact, physics-informed FC-NN methods have explored the parameterized spaces, including initial conditions, boundary conditions, and equation parameters, but FC-NN takes scalability problems reason of high-dimensional input spaces.
Because of parameter sharing and local connection, CNN is more promising in large-scale and high-dimensional problems.

In physics simulation field, the CNN have been applied in discoverying underlying PDEs discovery (\cite{long2018pde};\cite{long2019pde}), PDEs solving (\cite{yao2020fea};\cite{zhu2019physics}), and surrogate model construction (\cite{zhu2019physics}).
\cite{yao2020fea} presented that finite element method (FEA) models for PDEs is a special CNN, and FEA-Net was developed to predict the response of materials and structures.
\cite{kim2019deep} discretized and parameterized fluid simulation velocity fields, and a CNN-based generative model is provided to synthesize fluid simulations.
In our paper, the physics-informed CNN is utilized as surrogate model in thermal simulation task. The effect of different network components is thoroughly discussed, which is helpful for researches of network structure design in this field.  
In addition, for the HSL-TFP task at hand, considering different optimization difficulties in the computation domain, this paper proposes pixel-level online hard example mining to weigh each position adaptively. It can drive optimizer to make more efforts into hard-optimized domain.

In conclusion, the novel contributions of our work are as follows.
\begin{enumerate}
	\item [1)] We develop a physics-informed loss function for HSL-TFP based on heat conduction equation and finite difference method. The network can be trained by the physics-informed loss without labeled data.
	\item[2)] We introduce to impose hard constraints on the Dirichlet and Neumann boundary. This will benefit the stable and fast convergence of network training.
	\item[3)] A UNet-like architecture is proposed, and we emphasize discussing the performance impact of different network components.
	\item[4)] To balance the optimization difficulty in the whole computation domain, we introduce the pixel-level online hard example mining for HSL-TFP. The experiments demonstrate that the proposed method can present comparable predictions with numerical and data-driven methods.
\end{enumerate}

The rest of this paper is organized as follows. The description of the HSL-TFP problem is introduced in Section~\ref{sec:2}. Section~\ref{sec:3} presents the construction of physics-informed loss function, the hard constraints on boundary conditions, the CNN-based architecture, and the pixel-level hard example mining. The results of two cases with varying difficulty are presented and discussed in Section~\ref{sec:4}. Finally, we conclude the paper in Section~\ref{sec:5}.

\section{Temperature Field Prediction of Heat Source Layout (HSL-TFP)}
\label{sec:2}

\begin{figure}
	\centering
	\includegraphics[width=0.20\textheight]{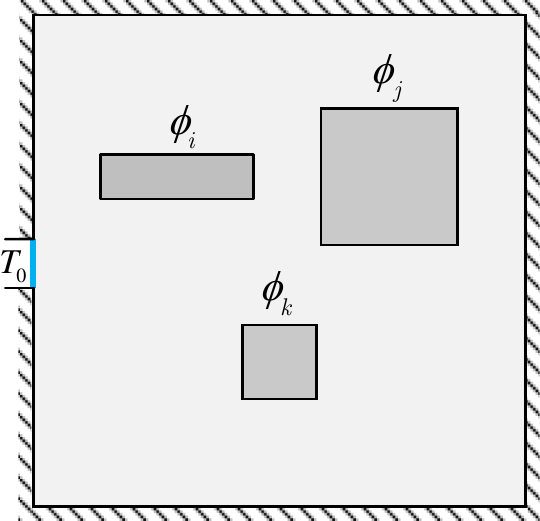}
	\caption{Schematic of the square domain with several heat source.}
	\label{fig:layout}
\end{figure}

Assume that there is a two-dimensional rectangular heat-conducting plate with several electronic components. The plate makes heat exchange with the external environment by isothermal boundary.
The components generate heat when working, and the heat is dissipated by conduction.
Each component can be simplified as a heat source for this type of problem, and the heat source layout will affect the steady temperature field distribution in the whole domain.
This paper expects to employ DNN to make temperature field predictions for heat source layout rapidly.

Similar to problems stated by \cite{chen2016,chen2017,chen2018,chen2020heat,chen2021deep}, the two-dimensional conducting domain with several rectangle heat source is set as shown in  Fig.~\ref{fig:layout}. An isothermal boundary of length $\delta$ and temperature $T_0$ in the bottom exchanges heat with external circumstances. Other boundaries are adiabatic. The steady temperature field of the domain satisfies Poisson's equation, which is expressed as follows.
\begin{equation}
\label{eq:1}
\frac{\partial }{{\partial x}}\left( {\lambda\frac{{\partial T}}{{\partial x}}} \right) + \frac{\partial }{{\partial y}}\left( {\lambda\frac{{\partial T}}{{\partial y}}} \right) + \phi (x,y) = 0, ~ (x, y) \in \Omega
\end{equation}
where constant $\lambda$ is the thermal conductivity of the domain. $\phi (x,y)$ is the intensity distribution function, which correlates with the heat source layout. Assume that ${\phi _i}$ and $\Gamma_i$ respectively represent the intensity and distribution area of the heat source $i$, $\phi (x,y)$ can be modeled as follows
\begin{equation}
\label{eq:2}
\phi (x,y) = \left\{ {\begin{array}{*{20}{c}}
	{{\phi _i},}&{(x,y) \in \Gamma_i ~~~ i \in 1,...,n}\\
	{0,}&{\rm{others}}
	\end{array}} \right.
\end{equation}
For the problem in this paper, the equation should satisfy the Drichlet Boundary and Neumann Boundary, which are formulated as follows 
\begin{equation}
\label{eq:3}
T(x,y) = {T_0}, (x, y) ~ \in \partial \Omega_D
\end{equation}
\begin{equation}
\label{eq:4}
\lambda\frac{{\partial T(x,y)}}{{\partial \bm{n}}} = 0, (x, y) ~ \in \partial \Omega_N
\end{equation}
where $\Omega_D$ and $\Omega_N$ respectively denote the Drichlet and Neumann Boundary. The equation can be solved by traditional numerical methods, such as finite difference method (FDM), finite element method (FEM), and finite volume method (FVM). Consequently, traditional numerical solvers sometimes are inefficient when calling the solver frequently. In this paper, we investigate to replace the numerical solver with neural networks to make real-time temperature field predictions.

\section{Physics-informed Convolutional Neural Networks for HSL-TFP}
\label{sec:3}

In this paper, each heat source layout corresponds to one PDE parameterized by the intensity distribution function $\phi (x,y)$. For the trained CNN, we expect the neural network takes $\phi (x,y)$ as input, and predict the two-dimentional temperature field $T$. Therefore, the network is essentially learning operators, mapping intensity distribution function to the PDE solution. 

In addition, some works have shown the power of deep networks in learning operators. Motivated by the success of deep learning, \cite{bhattacharya2021model} developed a data-driven framework to describe the mapping between infinite-dimensional spaces and applied it to the elliptic PDE. \cite{patel2021physics} introduced a regression framework to discover operators from molecular simulation data. \cite{anandkumar2020neural} and \cite{li2020multipole} proposed a graph kernel network to describe the mapping between different finite-dimensional approximations for infinite-dimensional spaces. By introducing Fourier transform, \cite{li2020fourier} presented the Fourier neural operator, which is a resolution-invariant solution operator for the Navier-Stokes equation. From the universal approximation theorem, \cite{lu2021learning} proposed a network, named deep operator network (DeepONet), which can make small generalization errors for deterministic and stochastic PDEs.

In contrast with these methods, the learning of deep operators is driven by physics equations in our work. For the HSL-TFP task, discrete representations of intensity distribution and solution functions are obtained by dividing mesh. These representations can be viewed as images, and CNN is employed for the image-to-image regression problem in finite dimension space. Guided by physics knowledge, CNN can learn to solve an entire family of heat conduction equations.

This section will first provide the construction of the physics-informed loss function for HSL-TFP, and the convolution layer can handily compute the loss. Then, we introduce the hard enforcement in prediction to satisfy the boundary condition. Finally, the CNN architecture based on UNet is described.

\subsection{Physics-informed loss function}

The Physics-informed CNN aims to learn the steady-state solutions of parametric PDEs described in Eq.~\ref{eq:1}.
Let $G$ be an operator taking the intensity distribution function $\phi$, and $G(\phi)$ is solution function of the corresponding PDE.
For each point $(x,y)$ in the solution $G(\phi)$, $G(\phi)(x,y)$ output the temperature, which is a real number.
We consider the two-dimensional heat conduction problem for HSL-TFP, and the computation domain is set to the rectangle. 
Without loss of generality, the rectangle domain and boundary are described as follows.
\begin{equation}
\label{eq:5}
D_I = \left\{ {(x,y)|0 < x < a,0 < y < b} \right\}
\end{equation}
\begin{equation}
\label{eq:6}
\begin{aligned}
\partial D =\{ (x,y)| 0 \le x \le a,y = 0{\rm{~or~}}b; x = 0 {\rm{~or~}} a,0 \le y \le b\}
\end{aligned}
\end{equation}
\begin{figure*}
	\centering
	\includegraphics[width=0.75\linewidth]{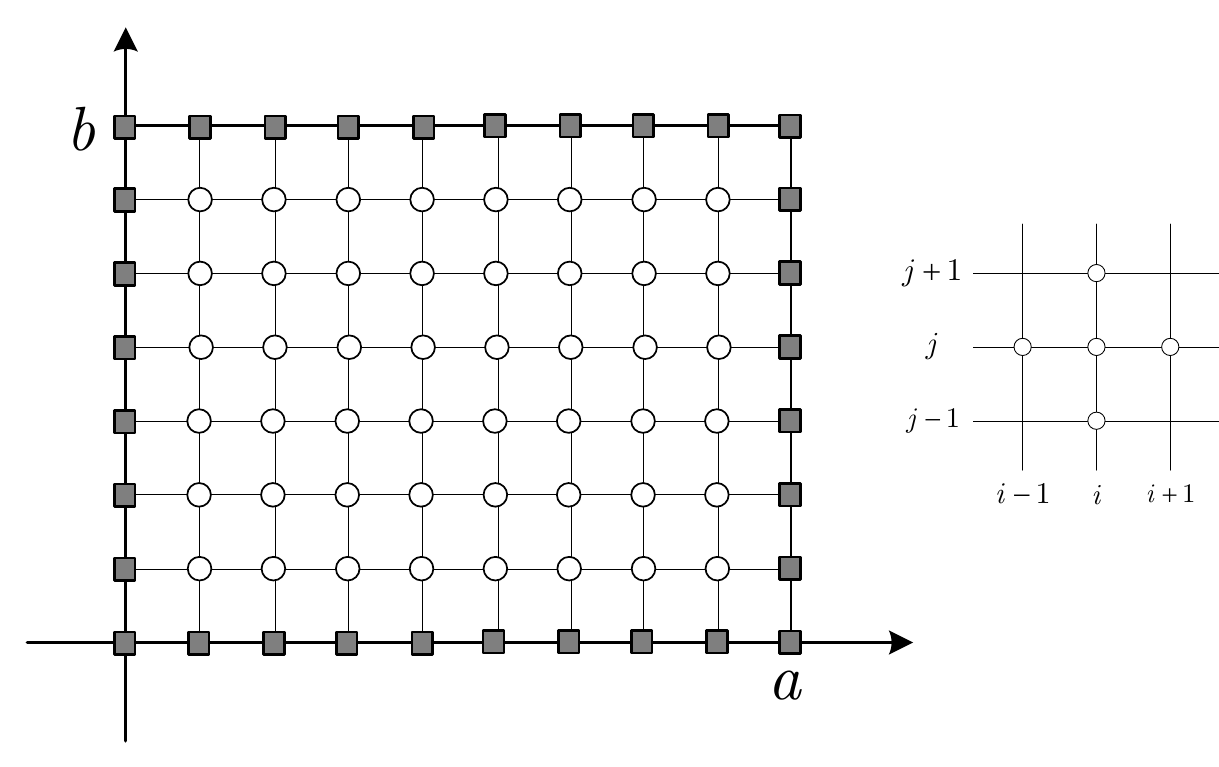}
	\caption{Illustration of mesh division in rectangle domain.}
	\label{fig:mesh}
\end{figure*}

To represent the function $\phi$ and $G(\phi)$, we discretize the function domain at finite amounts of location, and the function $\phi$ and $G(\phi)$ are treated as an image. Considering the great success of CNN in image processing, we expect that the operetor $G$ can be well learned by CNN.
As shown in Fig.~\ref{fig:mesh}, the computation domain $D$ bounded by $\partial D$ is divided into $n \times m$ rectangle mesh, and the step along the $x$-aixs and $y$-axis are $h = {a \mathord{\left/{\vphantom {a n}} \right.\kern-\nulldelimiterspace} n}$ and $k = {b \mathord{\left/{\vphantom {b m}} \right.\kern-\nulldelimiterspace} m}$.
Inspired by the finite difference method, the derivative terms of the solution function $G(\phi)$ could be obtained by differential approximation.
Therefore, we can construct a loss function to drive the solution function $G(\phi)$ satisfy the governing equation.
For inner-domain points $D_I$, the five-point finite difference is built by Taylor series expansion. 
\begin{equation}
\label{eq:7}
\begin{aligned}
\frac{1}{{{h^2}}}\left[ {T({x_i} + h,{y_j}) - 2T({x_i},{y_j}) + T({x_i} - h,{y_j})} \right] = \frac{{{\partial ^2}T}}{{\partial {x^2}}} + \frac{{{h^2}}}{{24}}\left[ {\frac{{{\partial ^4}}}{{\partial {x^4}}}T\left( {{\xi _1},{y_j}} \right) + \frac{{{\partial ^4}}}{{\partial {x^4}}}T\left( {{\xi _2},{y_j}} \right)} \right]
\end{aligned}
\end{equation}
where ${x_{i - 1}} \le {\xi _1},{\xi _2} \le {x_{i + 1}}$. 
similarly, the difference form of ${{{\partial ^2}T} \mathord{\left/{\vphantom {{{\partial ^2}T} {\partial {y^2}}}} \right.\kern-\nulldelimiterspace} {\partial {y^2}}}$ is presented as follows.
\begin{equation}
\label{eq:8}
\begin{aligned}
\frac{1}{{{k^2}}}\left[ {T({x_i},{y_j} + k) - 2T({x_i},{y_j}) + T({x_i},{y_j} - k)} \right] = \frac{{{\partial ^2}T}}{{\partial {y^2}}} + \frac{{{k^2}}}{{24}}\left[ {\frac{{{\partial ^4}}}{{\partial {y^4}}}T\left( {{x_i},{\eta _1}} \right) + \frac{{{\partial ^4}}}{{\partial {y^4}}}T\left( {{x_i},{\eta _2}} \right)} \right]
\end{aligned}
\end{equation}
where ${y_{i - 1}} \le {\eta _1},{\eta _2} \le {y_{i + 1}}$. According to Eq.~\ref{eq:7} and \ref{eq:8}, the difference equation of Eq.~\ref{eq:1} is obtained as follows.
\begin{equation}
\label{eq:9}
\begin{aligned}
& \frac{\partial }{{\partial x}}\left( {\lambda\frac{{\partial T}}{{\partial x}}} \right) + \frac{\partial }{{\partial y}}\left( {\lambda\frac{{\partial T}}{{\partial y}}} \right) + \phi (x,y) \\ = & \frac{\lambda}{{{h^2}}}\left[ {T({x_i} + h,{y_j}) - 2T({x_i},{y_j}) + T({x_i} - h,{y_j})} \right] \\ & + \frac{\lambda}{{{k^2}}}\left[ {T({x_i},{y_j} + k) - 2T({x_i},{y_j}) + T({x_i},{y_j} - k)} \right] + \phi (x,y) + O(h^2+k^2)
\end{aligned}
\end{equation}

The truncation error of the five-point finite difference is $O(h^2+k^2)$. For convenience of discussion, we use $x_{i - 1}$, $x_{i + 1}$ to represent $({x_i} - h)$ and $({x_i} + h)$, and the same is for y-axis. For each point $T(x_i,y_j)$ of mesh in $D_I$, it should satisfy the differential equation as follows.
\begin{equation}
\label{eq:10}
\begin{aligned}
\frac{{T\left( {{x_{i + 1}},{y_j}} \right) - 2T\left( {{x_i},{y_j}} \right) + T({x_{i - 1}},{y_j})}}{{{h^2}}} + \frac{{T\left( {{x_i},{y_{j + 1}}} \right) - 2T\left( {{x_i},{y_j}} \right) + T({x_i},{y_{j - 1}})}}{{{k^2}}} + \frac{\phi (x,y)}{\lambda} = 0
\end{aligned}
\end{equation}

The loss function can be constructed based on the above difference equation. This paper makes steps $h$ and $k$ equal in mesh generation for easily calculating by a convolution kernel. Under this setting, Eq.~\ref{eq:10} can be converted to the following form. 
\begin{equation}
\label{eq:11}
\begin{aligned}
4 \cdot T\left( {{x_i},{y_j}} \right) - T\left( {{x_{i + 1}},{y_j}} \right) - T({x_{i - 1}},{y_j}) - T\left( {{x_i},{y_{j + 1}}} \right) - T({x_i},{y_{j - 1}})  = \frac{{h^2} \phi (x_i,y_j)}{\lambda}
\end{aligned}
\end{equation}

\begin{figure}
	\centering
	\includegraphics[width=0.4\linewidth]{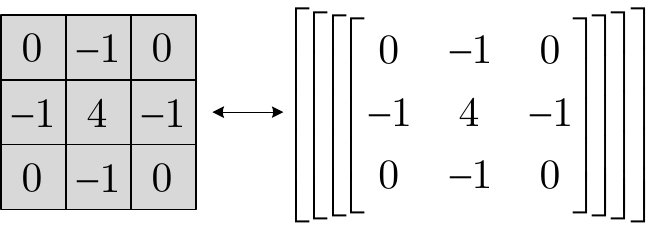}
	\caption{Laplace convolution kernel for second difference.}
	\label{fig:laplace}
\end{figure}
\begin{figure}
	\centering
	\includegraphics[width=0.4\linewidth]{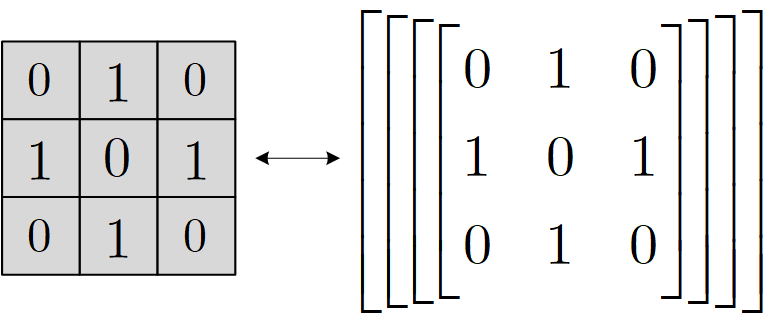}
	\caption{The convolution kernel for computing $T(x_i, y_j)'$.}
	\label{fig:laplace2}
\end{figure}
The temperature field $T$ is the solution $G(\phi)$ predicted by CNN. We expect each inner-domain point of $T$ to satisfy the above difference equation, and the second difference can be easily computed by the Laplace convolution kernel shown in Fig.~\ref{fig:laplace}. However, we discover that it is hard to train the neural network when roughly employing the Laplace convolution. Training is more efficient when the physics-informed loss function is constructed in the following form.
\begin{equation}
\label{eq:12}
\begin{aligned}
{\cal L} = \frac{1}{{\left| {{D_I}} \right|}}\sum\limits_{(x,y)} {\left\| {T\left( {{x_i},{y_j}} \right) - \frac{1}{4}T{{\left( {{x_i},{y_j}} \right)}^\prime }} \right\|}, (x,y) \in D_I
\end{aligned}
\end{equation}
\begin{equation}
\label{eq:13}
\begin{aligned}
T(x_i, y_j)' = T({x_{i + 1}},{y_j}) + T({x_{i - 1}},{y_j}) + T({x_i},{y_{j + 1}}) + T({x_i},{y_{j - 1}}) + \frac{{h^2}\phi (x,y)}{\lambda}
\end{aligned}
\end{equation}
where $T(x_i, y_j)'$ is only the intermediate variable. It can also be computed by the convolution kernel, which is shown in Fig~.\ref{fig:laplace2}. In particular, when computing $T(x_i, y_j)'$, the computation graph among $T({x_{i + 1}},{y_j})$, $T({x_{i - 1}},{y_j})$, $T({x_i},{y_{j + 1}})$, and $T({x_i},{y_{j - 1}})$ need not be constructed, which avoids that the error of $T(x_i, y_j)'$ affects neighboring points in back propagation. Although it is essentially derived from the difference equation of Eq.~\ref{eq:10}, this loss function formulation can significantly improve network training efficiency. In addition, the loss function can also be constructed by another difference scheme, such as the nine-point difference.

Why can the physics-informed loss function in this paper guide the neural network to learn the operator for solving an entire family of PDEs? It may be explained with the aid of the iteration method. Iterative numerical methods is an iterative algorithm to determine the solutions for the linear equations system. It is from an initial solution and bases on an iteration format to produce the approximation solution sequence. Jacobi method is a common iteration method, and the update rule for Poisson's equation is denoted in the following form.
\begin{equation}
\label{eq:14}
\begin{aligned}
T{\left( {{x_i},{y_j}} \right)^\prime } = \frac{ {T({{x_{i + 1}},{y_j}}) + T({x_{i - 1}},{y_j})} {+ T( {{x_i},{y_{j + 1}}}) + T({x_i},{y_{j - 1}})}}{4} + \frac{{{h^2}}}{4\lambda}\phi ({x_i},{y_j})
\end{aligned}
\end{equation}
The form of the loss function is consistent with Jacobi iteration. In each back propagation, the loss function guides the network to predict the temperature field approximating the value of the next iteration. Though the network learns to solve an entire family of PDEs, not one PDE, the iteration information of other PDEs is also helpful for training because of the similarity among parametric PDEs.  

\subsection{Hard-contrainted boundary condition}

\begin{figure*}
	\centering
	\includegraphics[width=1.0\linewidth]{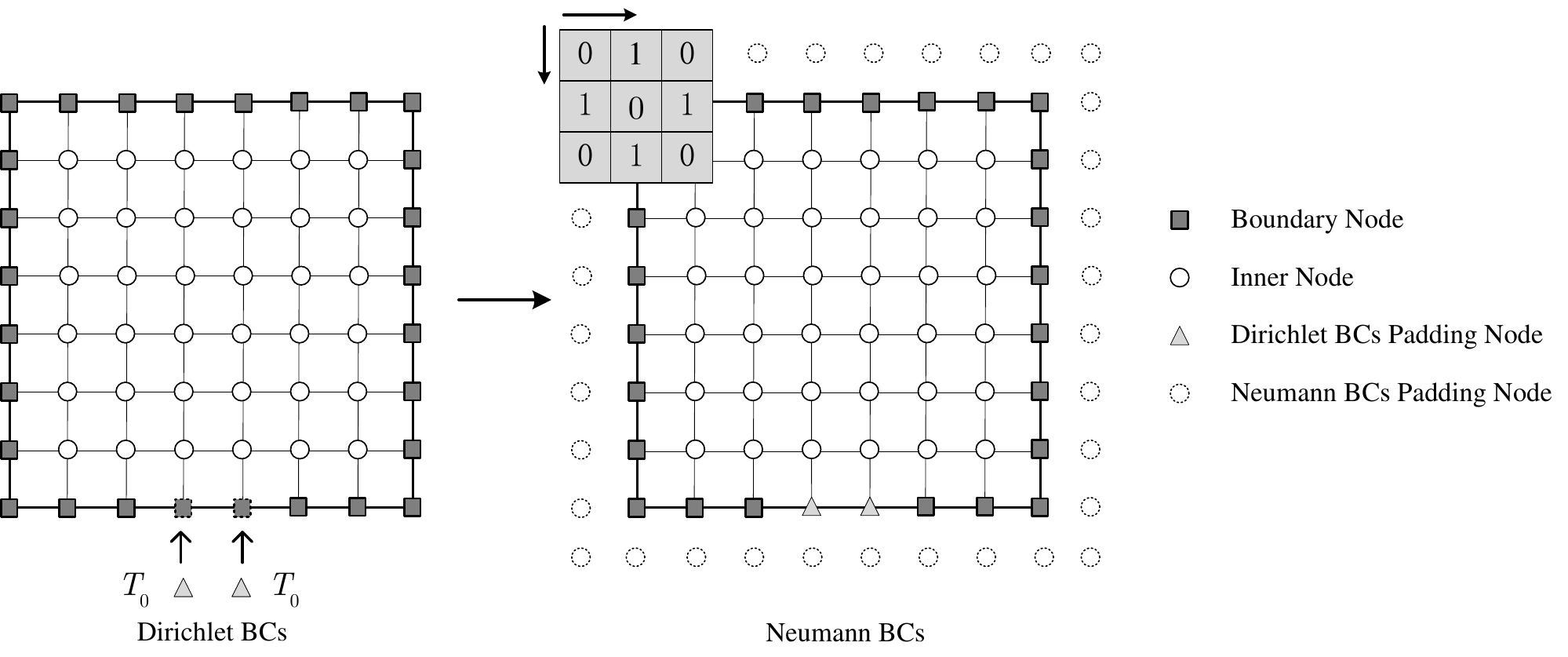}
	\caption{The illustration of enforcing hard constraint in Dirichlet and Neumann boundary.}
	\label{fig:boundary}
\end{figure*}
The boundary conditions uniquely determine the solution of the steady-state problem.
It is essential to add constraints on the computation domain properly. Otherwise, the PDE solving becomes ill-posed. 
One way is to apply soft constraints, which add penalty terms in the loss function to impel the neural network output specific values in the boundary. 
Since the solution is sensitive to the boundary values, the soft-constraint will negatively affect the early training when the boundary values do not fit exactly.
\cite{sun2020surrogate} introduced that soft boundary constraint will slow the convergence of PINN and devised a DNN structure to enforce the initial and boundary conditions. For CNN-based architecture, \cite{gao2021phygeonet} also studied the hard boundary condiction enforcement for solving parameterized PDE. 
To improve the stability and convergence speed, we also perform hard-constraint on the Dirichlet and Neumann boundary conditions for the HSL-TFP.

For the Dirichlet boundary condition defined in Eq.~\ref{eq:3}, the values in Dicichlet boundary are directly replaced by the constant value $T_0$, which enforces that these values do not vary during training. 
For the Neumann boundary condition (Eq.~\ref{eq:4})), it is assumed that there is extra mesh $\partial \Omega '$ surrounding the computation domain $\Omega  \cup \partial \Omega$. According to the central difference formula, it should satisfy the following formulation.
\begin{equation}
\label{eq15}
\left\{
\begin{aligned}
& \lambda \frac{{{T(x_{n + 1}, y_j)} - {T(x_{n - 1}, j)}}}{{2h}} = 0 , & 0 \le j \le m, \\
& \lambda \frac{{{T(x_{-1}, y_j)} - {T(x_1, y_j)}}}{{2h}} = 0 , & 0 \le j \le m, \\
& \lambda \frac{{{T(x_i, y_{m + 1})} - {T(x_i, y_{m -1})}}}{{2h}} = 0 , & - 1 \le i \le n + 1, \\
& \lambda \frac{{{T(x_i, y_{-1})} - {T(x_i, y_1)}}}{{2h}} = 0 , & - 1 \le i \le n + 1. 
\end{aligned}
\right.
\end{equation}
Therefore, we can enforce the padding values equal to the corresponding inner-nodes.
\begin{equation}
\label{eq:16}
\left\{
\begin{aligned}
& T(x_{n + 1}, y_j) = T(x_{n - 1}, y_j), & 0 \le j \le m, \\
& T(x_{-1}, y_j) = T(x_1, y_j), & 0 \le j \le m, \\
& T(x_i, y_{m + 1}) = T(x_i, y_{m - 1}), & - 1 \le i \le n + 1, \\
& T(x_i, y_{-1}) = T(x_i, y_1), & - 1 \le i \le n + 1. 
\end{aligned}
\right.
\end{equation}
The schematic of enforcing hard-constrain in Dirichlet and Neumann boundary is shown in Fig.~\ref{fig:boundary}. Before computing the loss for gradient-based optimization, the prediction is padding with constant boundary values for Dirichlet BCs. Then the surrounding domain is expanded with inner-domain nodes for Neumann BCs. Finally, the processing map is applied to obtain the $T'$ according to Eq.~\ref{eq:12}, and the loss is calculated without the values in Dirichlet BCs. 
\begin{equation}
\label{eq:17}
\begin{aligned}
\mathcal{L} = \frac{1}{{\left| {{D_I} \cup {D_N}} \right|}} \sum\limits_{(x,y)} {\left\| {T({x_i},{y_j}) - T({x_i},{y_j})'} \right\|}, ({x_i},{y_j}) \in ({D_I} \cup {D_N})
\end{aligned}
\end{equation}

\subsection{CNN architecture and framework for HSL-TFP}

\begin{figure*}
	\centering
	\includegraphics[width=1.0\linewidth]{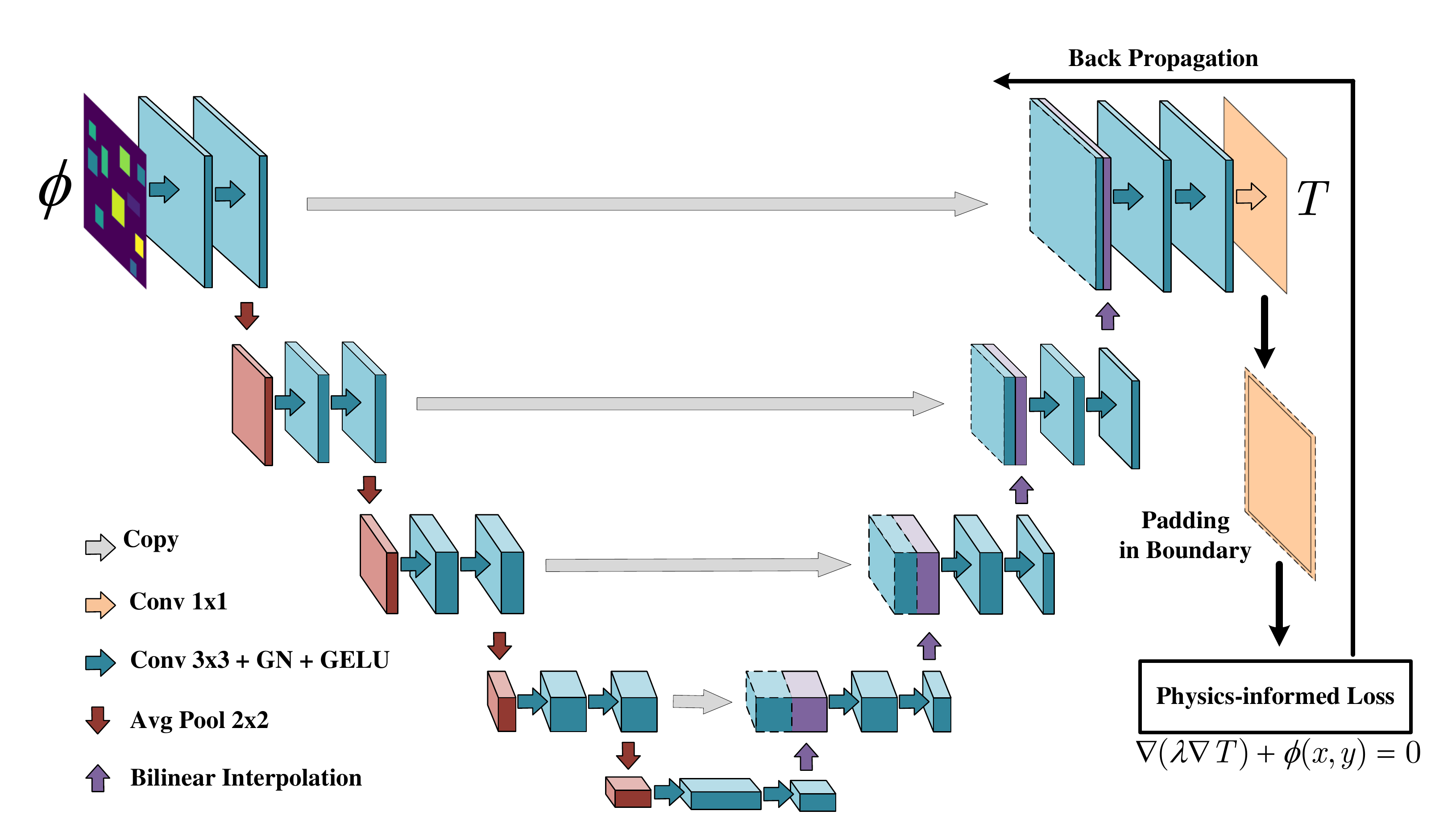}
	\caption{The pipline of physics-informed CNN for HSL-TFP. The CNN architecture is modified based on UNet for this specific task.}
	\label{fig:unet}
\end{figure*}

The inputs and outputs are both gridding data when applying the finite difference method for the two-dimensional steady-state prediction problem. 
When flattening the data, FC-NN will suffer from scalability problems reason of high-dimensional data space. 
Compared with FC-NN, CNN is superior to process image-style data.  
In this paper, the UNet (\cite{ronneberger2015u}) is employed as the base model, which is one excellent image-to-image CNN architecture. 
UNet is an encoder-decoder framework consisting of a contracting path to capture global information and an asymmetric expanding path to make high-resolution predictions. In particular, there is a skip connection in the same level of feature map for feature fusion from the encoder to decoder. Due to these special structures, UNet has achieved great success in many computer vision tasks.

Different from classification problems, the task at hand is an image-to-image regression problem. We modified UNet architecture, which is shown in Fig.~\ref{fig:unet}. 
The inputs and outputs are respectively the discretized representation of the source term $\phi(x,y)$ and $T$ in Eq.~\ref{eq:1}. 
The blue arrow is the block with the 3x3 convolution layer, GroupNorm (GN), and GELU (\cite{ronneberger2015u}) activation function. We choose GN (\cite{wu2018group}) because it is more stable than BacthNorm (BN) in this task. Since the values of intensity without heat source are zeros, the intensity distribution map is sparse. It changes drastically along the edge of the heat source, which is significantly different from visual images. The statistics of BN would not be exact and will harm training. 
In addition, GELU activation is expected to increase the ability in the nonlinear fitting.
The red downward arrow is the downsampling block using average pool with 2x2 filters and stride 2. Symmetrically, the upward arrow is the 2x upsampling block with bilinear interpolation rather than the transpose convolution.
Since the architecture is symmetrical, the padding of the 3x3 convolution layer is modified to 1 in contrast with original UNet, which makes the sizes of input and output under the same grid division are equal.
Besides, it is worth noticing that the padding mode of convolution layers are all set to 'reflect'.

When conducting network training, we discovered large differences in the optimization difficulty of different regions. The difference equations connect the points in the physics-informed loss function, which means errors in these regions will propagate to others. To address this issue, we propose pixel-level online hard example mining (P-OHEM) to balance the optimization dynamically . The error in each pixel of the output map is represented as
\begin{equation}
\label{eq:18}
{\delta _{ij}} = \left\| {T({x_i},{y_j}) - T({x_i},{y_j})'} \right\|
\end{equation}
Making $\delta$ the set of pixel-level error in map, the weight of pixel in position $(i, j)$ can be denoted as
\begin{equation}
\label{eq:19}
{w_{ij}} = {\eta _1} + {\eta _2}\frac{{{\delta _{ij}} - \max (\delta )}}{{\max (\delta ) + \min (\delta )}}
\end{equation}
where $\eta _1$ and $\eta _2$ are the shift and scale factors. Therefore, the loss function with P-OHEM is expresses as follows.
\begin{equation}
\label{eq:20}
\begin{aligned}
\mathcal{L} = \frac{1}{{\left| {{D_I} \cup {D_N}} \right|}} \sum\limits_{(x,y)} w_{ij}{\left\|{T({x_i},{y_j}) - T({x_i},{y_j})'} \right\|}, ({x_i},{y_j}) \in ({D_I} \cup {D_N})
\end{aligned}
\end{equation}

\begin{algorithm}[t]
	\caption{The framework of the physics-informed convolutional neural networks for HSL-TFP}
	\hspace*{0.02in} {\bf Input:}
	Intensity distribution set $\Phi$, Network parameters $\theta$, Hyperparameter $\eta_1$,$\eta_2$. \\
	\hspace*{0.02in} {\bf Output:}
	${\theta ^ * }$
	\begin{algorithmic}[1]
		\State Randomly initialize $\theta$
		\While{not done}
		\State Sample batch of intensity distribution $\Phi_N$ from $\Phi$
		\State Predict the temperature field $T_N$ with $\theta$
		\State Process boundary of $T_N$ with hard-constraint
		\State Calculate ${T_N}^\prime$ with Eq.~\ref{eq:13} in no auto-grad mode
		\State Compute pixel-level weight $w_{ij}$ with Eq.~\ref{eq:18} and Eq.~\ref{eq:19}
		\State Compute the physics-informed loss $\mathcal{L}$ with Eq.~\ref{eq:20}
		\State Update $\theta$ using $\mathcal{L}$ by auto-grad 
		\EndWhile
		\State \textbf{end while} 
		\State \Return optimized prameters ${\theta ^ * }$
	\end{algorithmic}
\end{algorithm}

For each sample, P-OHEM makes the optimizer more effort on the position with larger error, and the weight changes dynamically with the training. Benefit from this clear idea, we expect the error of each position can be uniformly decreased. Finally, the training pseudocode of the physics-informed convolutional neural networks for HSL-TFP is given in Alogorithm~1.

\section{Experiment}
\label{sec:4}

\subsection{Datasets and experimental settings}

\label{sec:4.1}

\begin{figure*}[tp]
	\centering
	\subfigure[layout of case 1]{
		\includegraphics[width=0.35\linewidth]{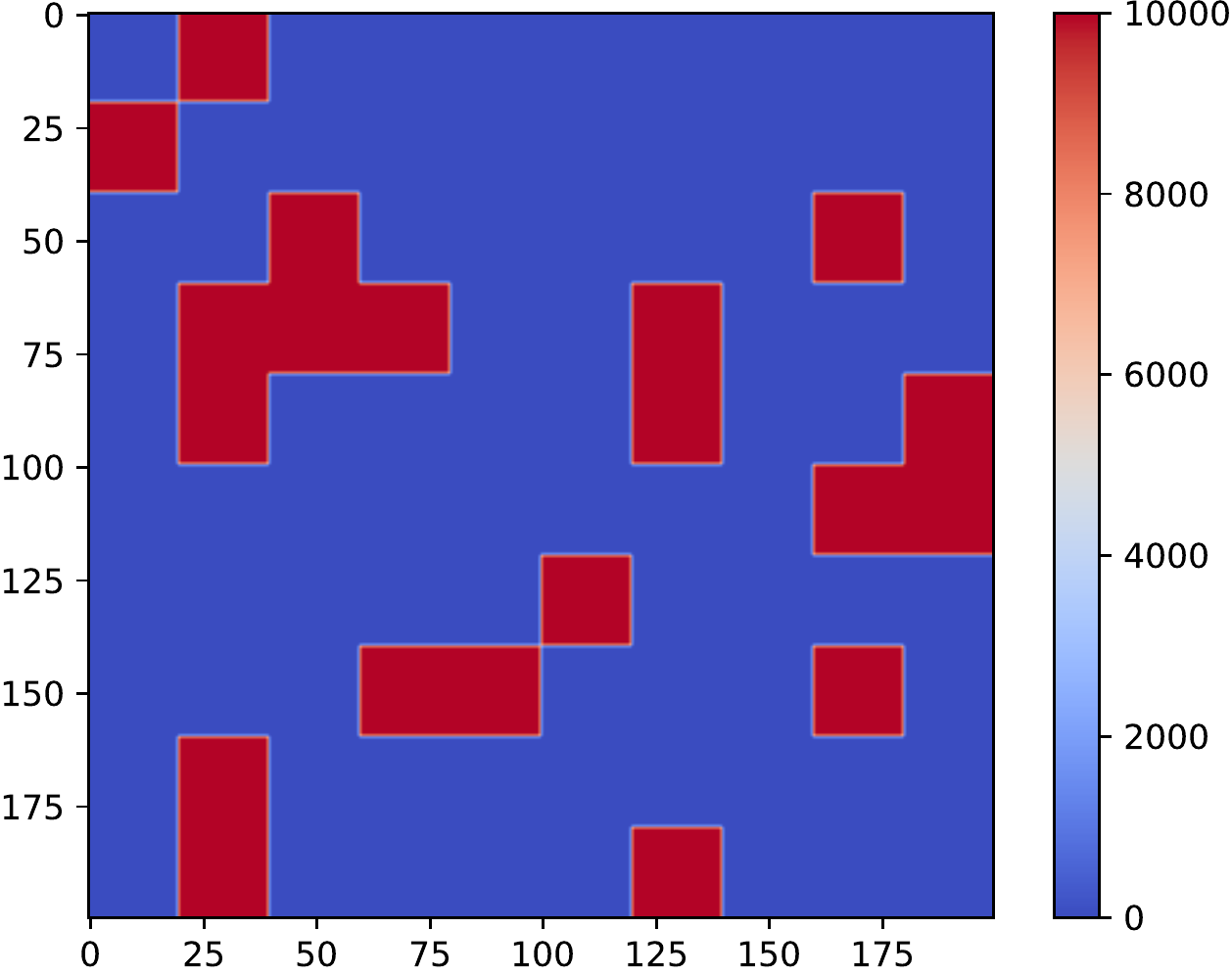}
	}
	\subfigure[temperature field of case 1]{
		\includegraphics[width=0.37\linewidth]{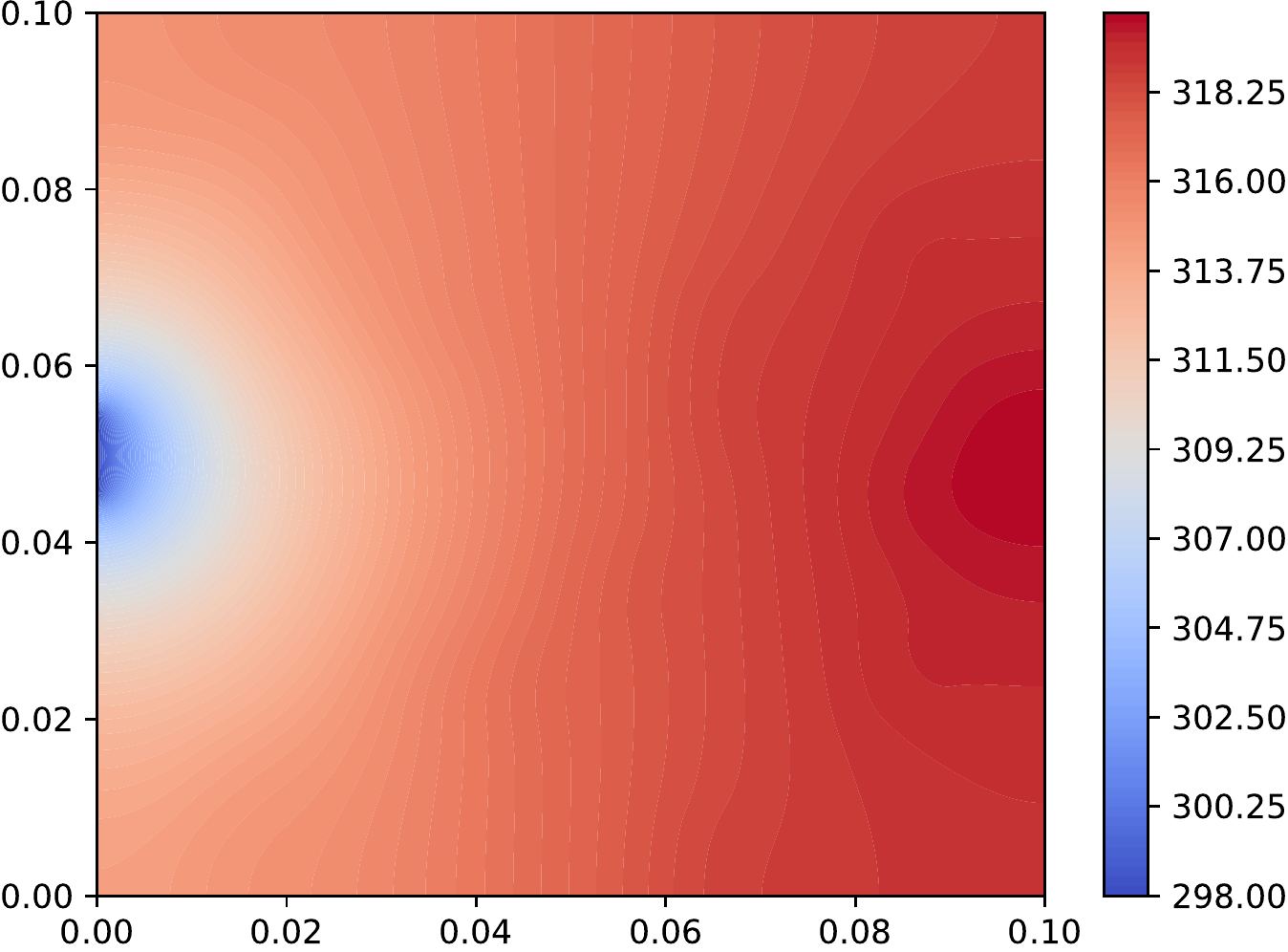}
	}
	\\
	\subfigure[layout of case 2]{
		\includegraphics[width=0.35\linewidth]{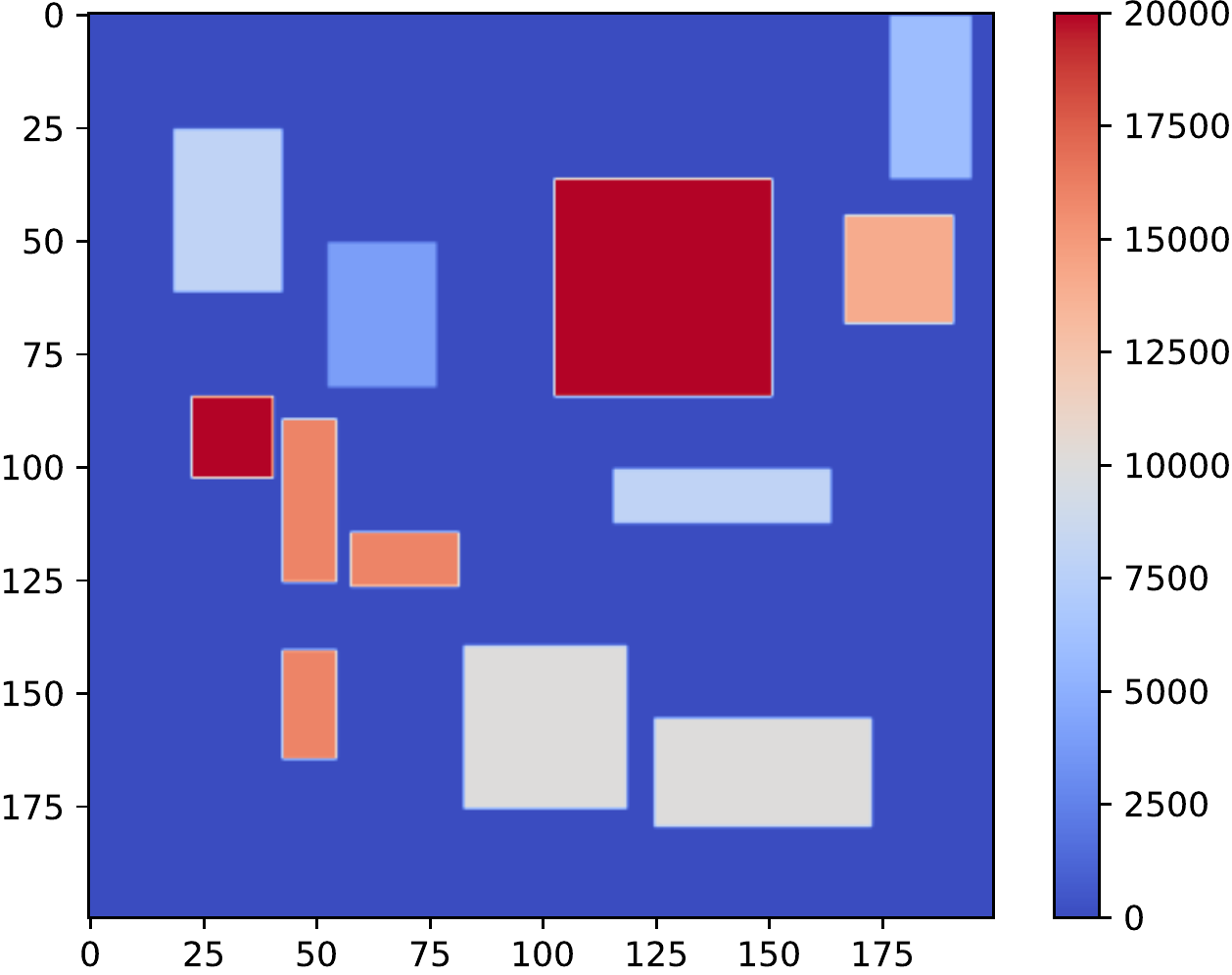}
	}
	\subfigure[temperature field of case 2]{
		\includegraphics[width=0.37\linewidth]{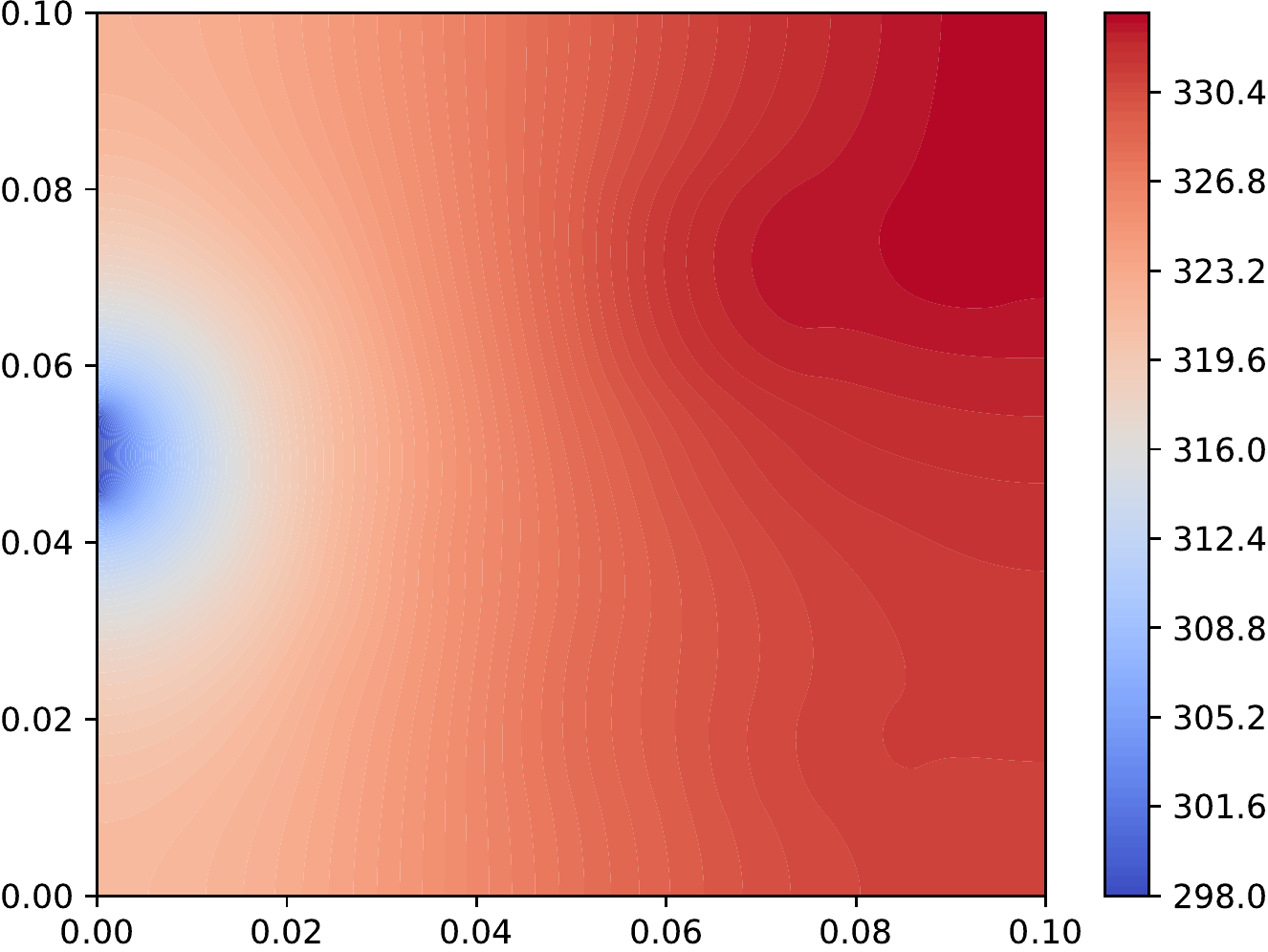}
	}
	\caption{The layout and temperature field of two cases.}
	\label{fig:cases}
\end{figure*}

\begin{table*}[htbp]
	\centering
	\caption{The configuration of heat source in case 1.}
	\begin{tabular}{ccc}
		\hline\noalign{\smallskip}
		Component & Size(m $\times$ m) & Intensity($w/m^2$) \\
		\noalign{\smallskip}\hline\noalign{\smallskip}
		1-20  & 0.01 $\times$ 0.01 & 10000 \\
		\noalign{\smallskip}\hline
	\end{tabular}%
	\label{tab:simple_confi}%
\end{table*}%

\begin{table*}[htbp]
	\centering
	\caption{The configuration of heat source in case 2.}
	\begin{tabular}{ccc}
		\hline\noalign{\smallskip}
		Component & Size($m \times m$)   & Intensity($w/m^2$)       \\
		\noalign{\smallskip}\hline\noalign{\smallskip}
		1         & 0.016 $\times$ 0.012 & 4000                     \\
		2         & 0.012 $\times$ 0.006 & 16000                    \\
		3         & 0.018 $\times$ 0.009 & 6000                     \\
		4         & 0.018 $\times$ 0.012 & 8000                     \\
		5         & 0.018 $\times$ 0.018 & 10000                    \\
		6         & 0.012 $\times$ 0.012 & 14000                    \\
		7         & 0.018 $\times$ 0.006 & 16000                    \\
		8         & 0.009 $\times$ 0.009 & 20000                    \\
		9         & 0.006 $\times$ 0.024 & 8000                     \\
		10        & 0.006 $\times$ 0.012 & 16000                    \\
		11        & 0.012 $\times$ 0.024 & 10000                    \\
		12        & 0.024 $\times$ 0.024 & 20000                    \\ \noalign{\smallskip}\hline
	\end{tabular}
	\label{tab:complex_confi}%
\end{table*}

\noindent \textbf{Datasets.} 
In this section, two cases with varying degrees of complexity are studied to demonstrate the proposed method. 
Firstly, the parameters of two cases are presented. The conduction domain is set to 0.1m $\times$ 0.1m square-shaped areas. One isothermal sink with a length of 0.01m is located in the middle of the left boundary, and the other is adiabatic. 
The thermal conductivity $\lambda$ of both domain and heat source equals $1W/(m \cdot K)$. 
For case 1, several square-shaped heat sources with the same setting are placed in the computation domain. For the more complex case, there are 12 rectangular heat sources with various sizes and intensities. The configurations of heat source for two cases are respectively presented in Table~\ref{tab:simple_confi} and Table~\ref{tab:complex_confi}.

\noindent \textbf{Evaluation Metrics.} \cite{chen2021deep} proposed a variety of metrics to evaluate the neural network performance for temperature field prediction. In this paper, we choose the following metrics, including the mean absolute error on the whole domain (MAE), the mean absolute error on the domain with components (CMAE), the maximum absolute error on the whole domain (Max-AE), and the absolute error of the maximum temperature (MT-AE).

\noindent \textbf{Implementation Details.} 
For each case, we sample 10000 heat source layout maps, and the sizes of training, validation, and test set are 9000, 1000, and 1000. The layout sampling methods are consistent with \cite{chen2020heat} and \cite{chen2021deep}. The hyperparameters of ${\eta _1}$ and ${\eta _2}$ are set to 0 and 10, respectively. For training, we utilize Adam optimizer with the initial learning rate of 0.01, and the learning rate is adjusted by polynomial decay policy with 0.85 multiplicative factor. The batch size and the total epoch are set to 1 and 30. 
All models are trained on a single NVIDIA GeForce RTX 3090 GPU. The Pytorch implementation is released at \url{https://github.com/zhaoxiaoyu1995/PI-UNet_HSL-TFP}.

\subsection{Prediction Performance}

\begin{table}[htbp]
	\centering
	\caption{The comparison of FDM computations and PI-UNet predictions.}
	\begin{tabular}{ccrrrcrrr}
		\hline\noalign{\smallskip}
		\multirow{2}[0]{*}{} & \multicolumn{4}{c}{Training Set} & \multicolumn{4}{c}{Test Set} \\
		& MAE   & \multicolumn{1}{c}{CMAE} & \multicolumn{1}{c}{Max-AE} & \multicolumn{1}{c}{MT-AE} & MAE   & \multicolumn{1}{c}{CMAE} & \multicolumn{1}{c}{Max-AE} & \multicolumn{1}{c}{MT-AE} \\
		\noalign{\smallskip}\hline\noalign{\smallskip}
		Case 1 & 0.0098 & 0.0099 & 0.0235 & 0.0123 & 0.0108 & 0.0109 & 0.0280 & 0.0140 \\
		Case 2 & 0.0230 & 0.0228 & 0.0536 & 0.0277 & 0.0275 & 0.0271 & 0.0925 & 0.0336 \\
		\noalign{\smallskip}\hline
	\end{tabular}%
	\label{tab:FDMandUL}%
\end{table}%

\begin{figure*}
	\centering
	\includegraphics[width=0.95\linewidth]{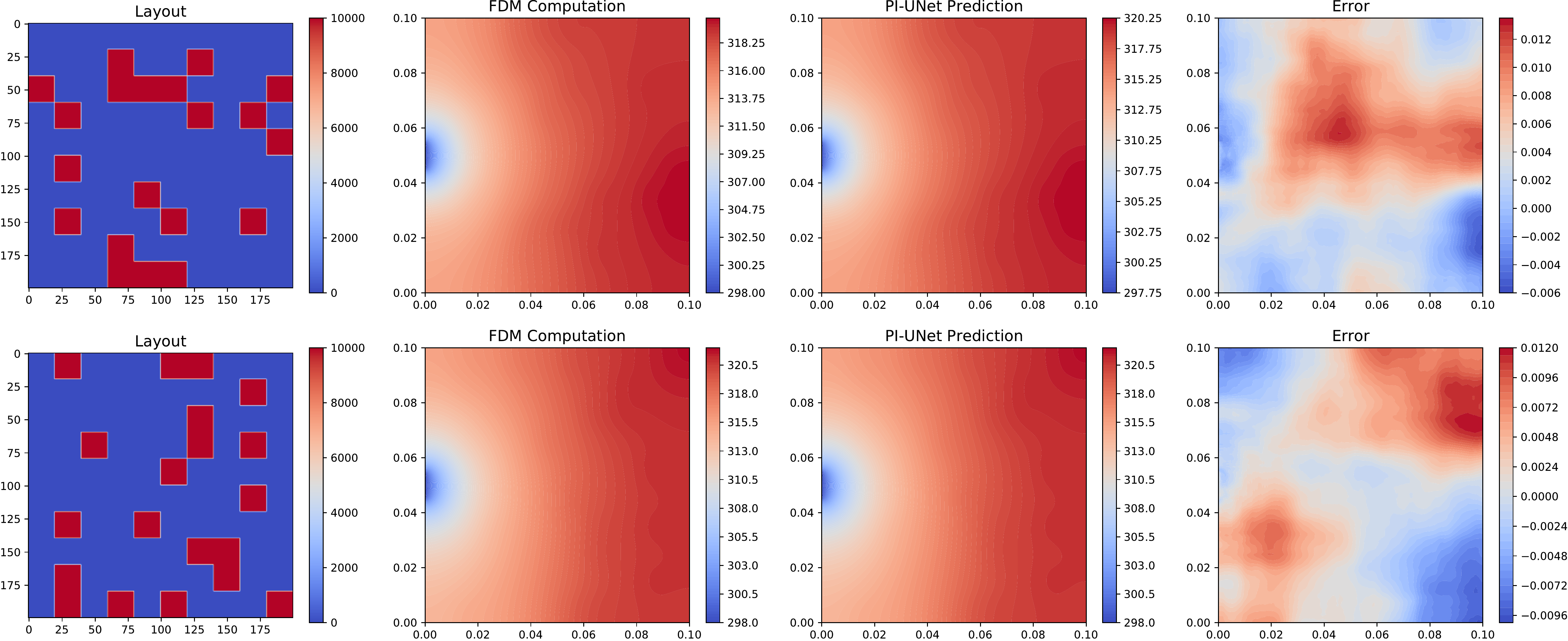}
	\caption{Prediction examples of case 1.}
	\label{fig:case1_pre}
\end{figure*}

\begin{figure*}
	\centering
	\includegraphics[width=0.95\linewidth]{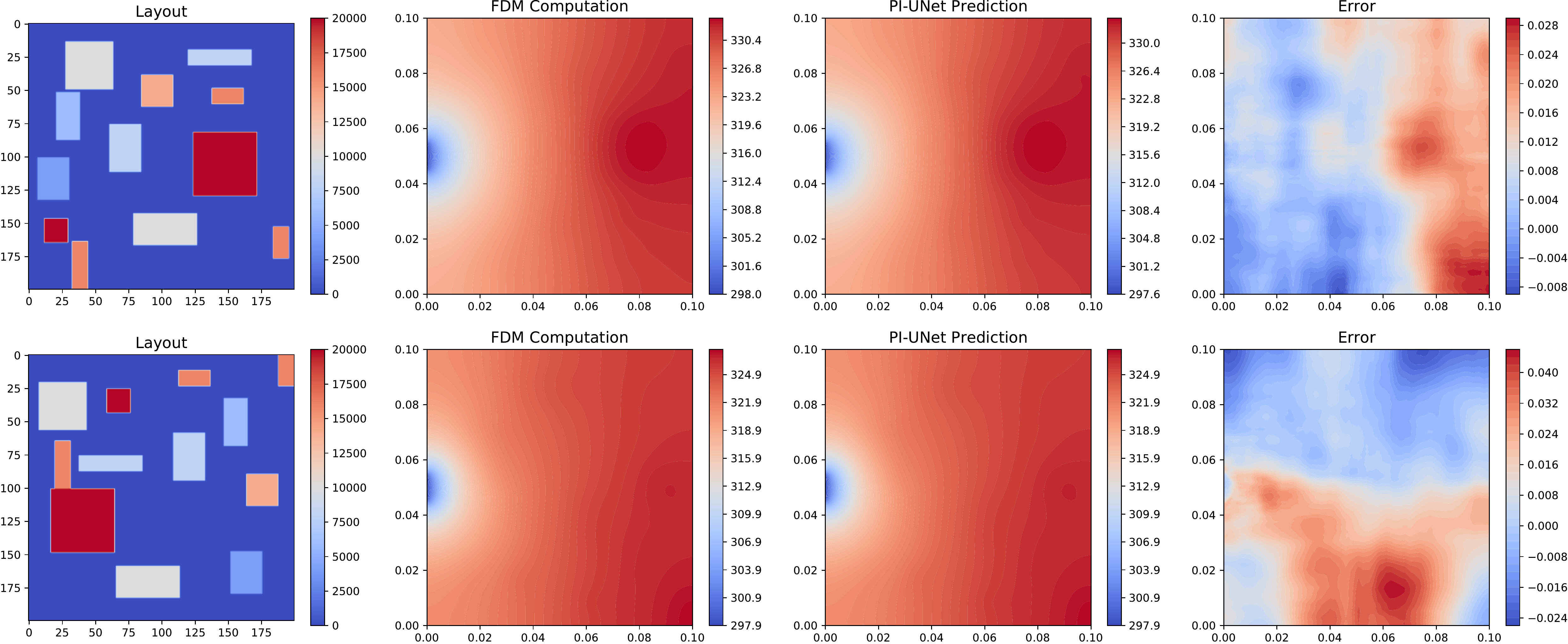}
	\caption{Prediction examples of case 2.}
	\label{fig:case2_pre}
\end{figure*}

Numerical method is sophisticatedly to compute the temperature field by solving PDEs. In this paper, finite difference method (FDM), one of the traditional numerical methods, is chosen as the benchmark to verify the proposed method. The implementation of FDM is following \cite{reimer2013matlab}.

As the physics-informed loss guides the CNN training without labeled data, we evaluate the network performance in the training and test sets. Our method is called as PI-UNet, and the results are presented in Table~\ref{tab:FDMandUL}. 
First, the results show PI-UNet can make high precision predictions, as the MAE with FDM computation is less than 0.03K in two cases. Case 2 is more difficult than case 1 because all evaluation metrics are larger. 
Second, the difference between various evaluation metrics is small. In particular, Max-AE evaluates the maximum absolute error, which is smaller than 0.1K. It means the error between prediction and FDM computation is uniform. It is essential because unusual errors occurring in the local region will damage the prediction reliability.
Finally, the small gap of evaluation metrics in the training and test set indicates that the proposed physics-informed CNN is well generalized. 
Fig.~\ref{fig:case1_pre} and Fig.~\ref{fig:case2_pre} respectively show the visual temperature field examples of prediction and FDM computation in case 1 and case 2. The network prediction is great close to the FDM computation. 
Since the prediction of the trained network is end-to-end and the forward propagation is very fast, the prediction by the trained network is superior in time consumption. In conclusion, the proposed method can guarantee rapid and high-precision prediction.

\subsection{Comparisons with supervised learning}

\begin{figure*}[tp]
	\centering
	\subfigure[case 1]{
		\includegraphics[width=0.40\linewidth]{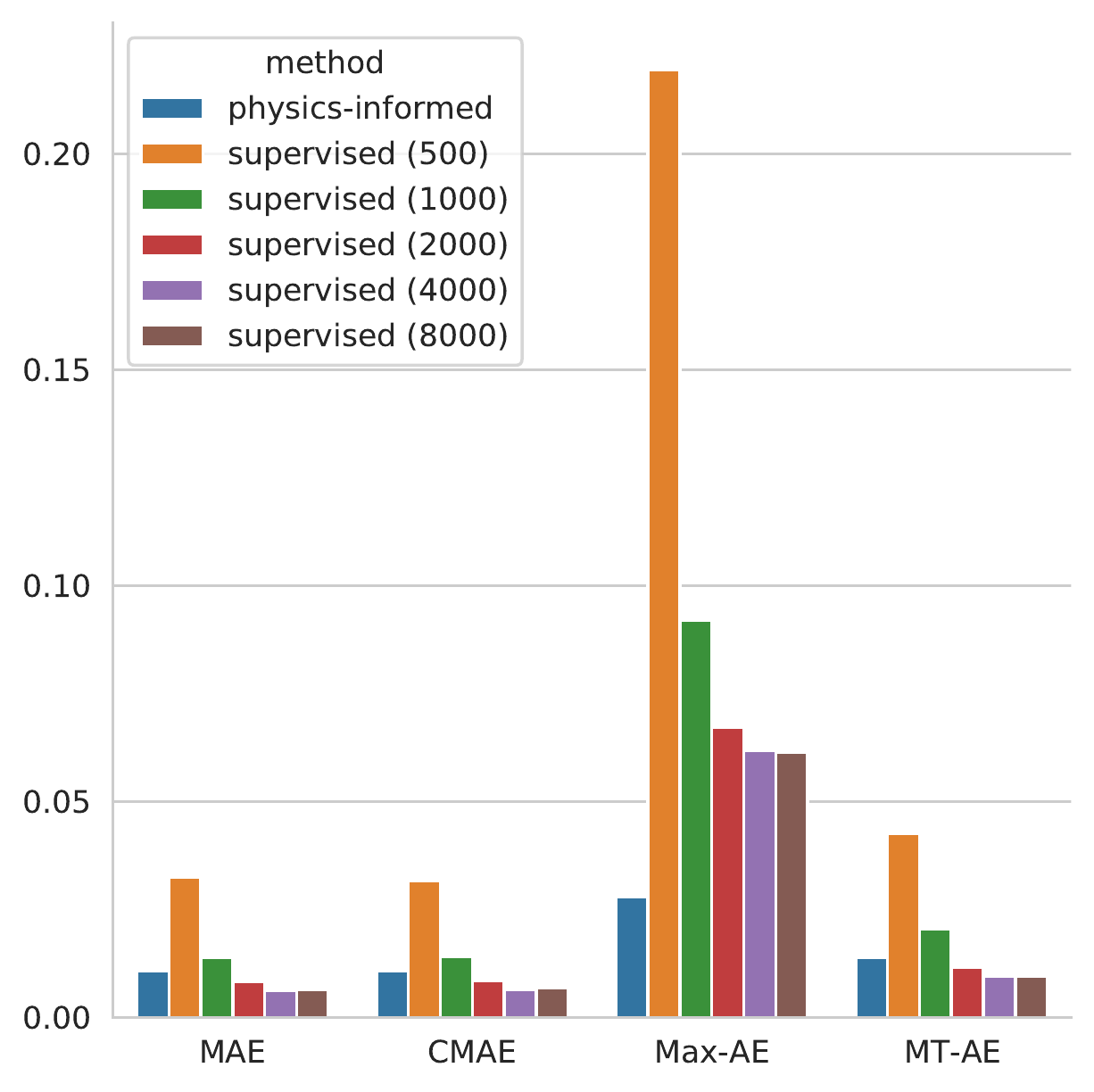}
	}
	\subfigure[case 2]{
		\includegraphics[width=0.40\linewidth]{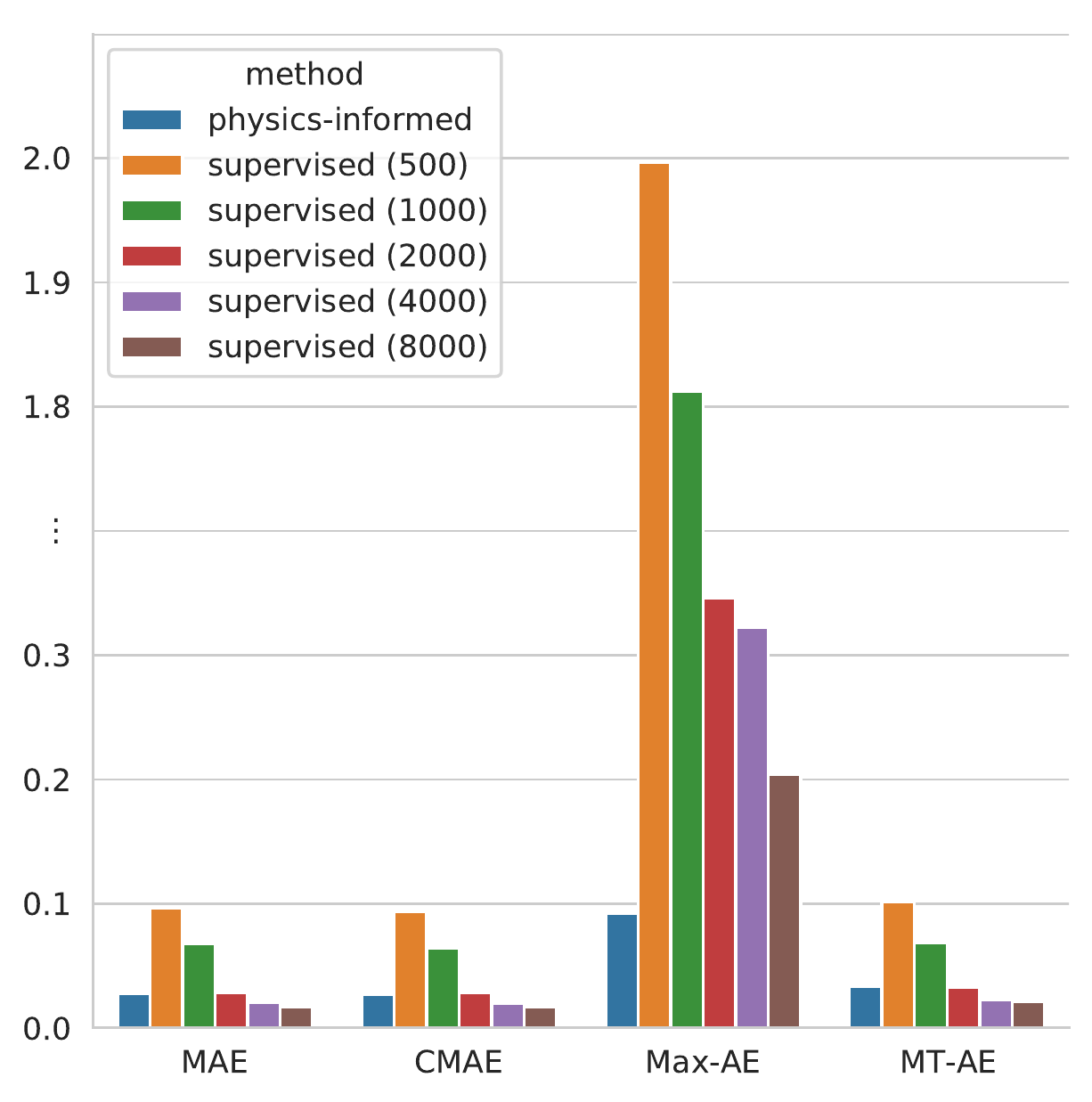}
	}
	\caption{Results of physics-informed learning and supervised learning with varying amounts of training data.}
	\label{fig:ul_sl}
\end{figure*}

The proposed method employs physics knowledge to guide the network training. To investigate physics-informed learning and traditional supervised learning characteristics for HSL-TFP,  we conduct supervised learning experiments with varying amounts of labeled data. The experimental setup is the same as physics-informed learning, except the physics-informed loss function is not applied. The number of labeled data is respectively set to 500, 1000, 2000, 4000, and 8000.

The results are compared in Fig.~\ref{fig:ul_sl}. From MAE, CMAE, and MT-AE, the precision of physics-informed learning is close to the supervised learning with 4000 labeled data. 
With the increase of labeled data, the model accuracy of supervised learning is higher, but there is no significant improvement.
Supervised learning performs better when the labeled data is more than 4000.
However, the gap between physics-informed learning and supervised learning is slight, and the precision of the proposed method meets the requirements of the application.
For Max-AE, physics-informed learning is significantly lower than supervised learning. 
Max-AE evaluates the maximum absolute error with the FDM computation. It indicates that the supervised paradigm only learns to approximate the whole domain from labeled data, and the prediction precision in the local region is weaker than the guidance of physics. It is promising to conduct semi-supervised learning to combine the physics guidance and labeled data.

\subsection{Ablation Studies}

We conduct a set of ablation experiments to study the effect of the loss function, training skills, and several network components.

\begin{figure*}[tp]
	\centering
	\subfigure[case 1]{
		\includegraphics[width=0.35\linewidth]{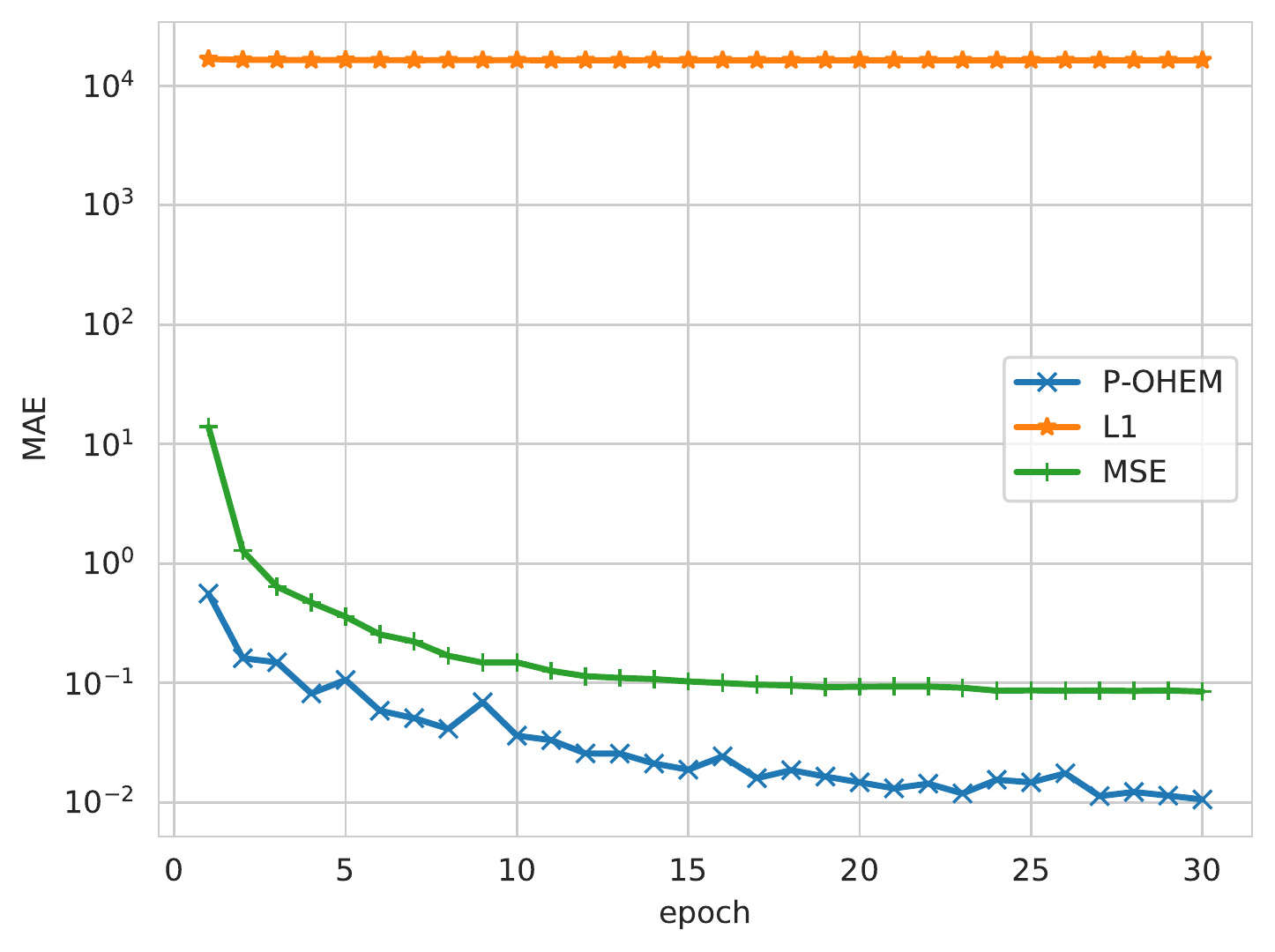}
	}
	\subfigure[case 2]{
		\includegraphics[width=0.35\linewidth]{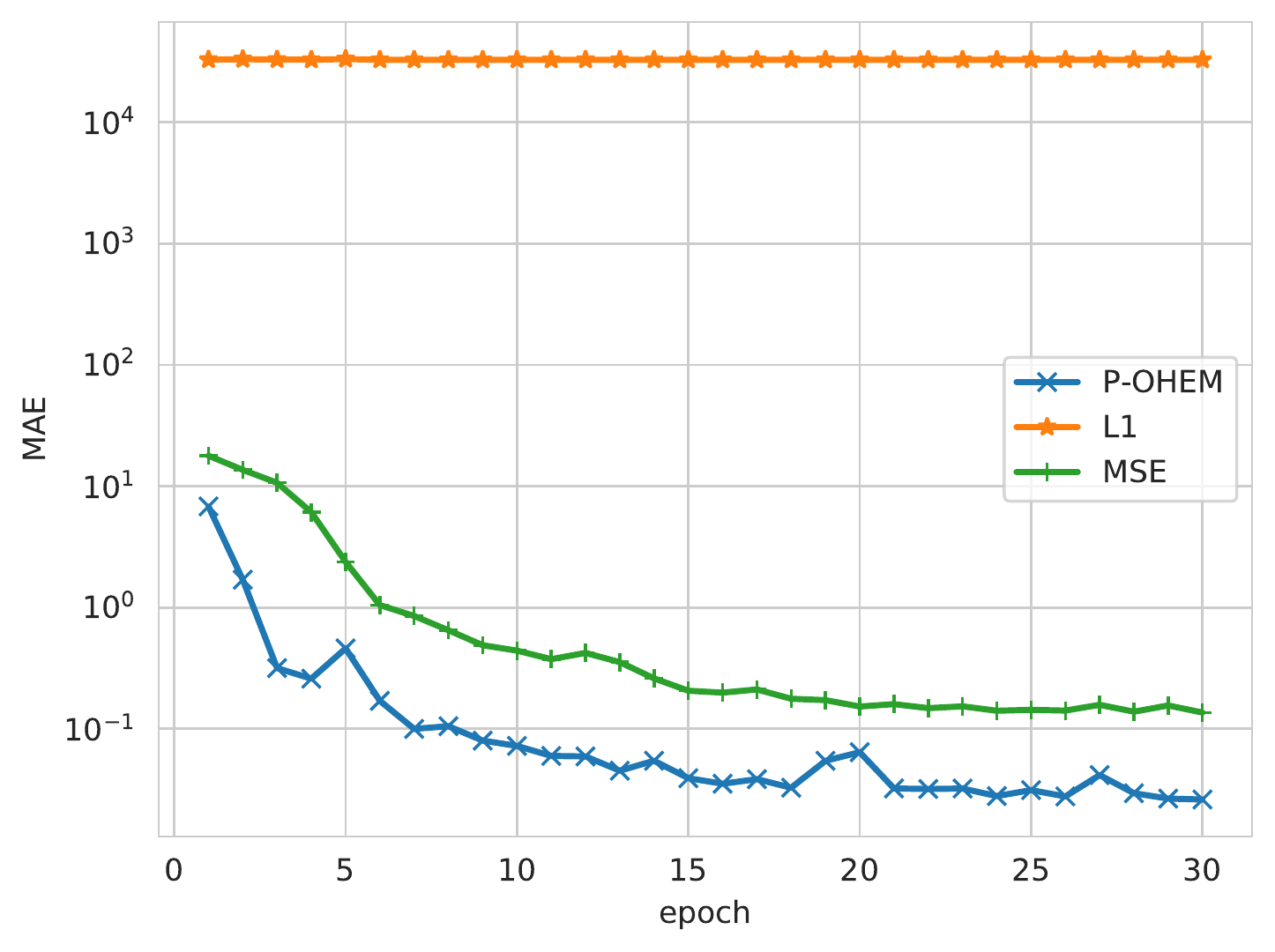}
	}
	\caption{The trend of validation MAE trained with O-OHEM, L1, and MSE loss.}
	\label{fig:ohem}
\end{figure*}

\noindent \textbf{Effect of P-OHEM.} We perform ablation experiments to compare the performance of L1, MSE, and the proposed P-OHEM loss functions. 
The results of MAE evaluated in test data after each epoch can be seen in Fig.~\ref{fig:ohem}. 
When employing L1 loss, the network is unable to optimize properly, but this issue does not occur when using MSE and P-OHEM loss. 
Since the physics-informed loss is constructed based on PDEs and finite difference, numerical stability may be affected by the L1 loss. 
Result from different optimization difficulties in the computation domain, P-OHEM performs much better than MSE. 
In particular, since the heat dissipated from the isothermal hole and temperature changes dramatically in the domain close to the hole, decreasing the loss in these domains is more challenging. 
In P-OHEM, there is a larger weight for the position with a larger error, and more efforts will be made to decrease the physics-informed loss. 

\begin{figure*}
	\centering
	\includegraphics[width=0.95\linewidth]{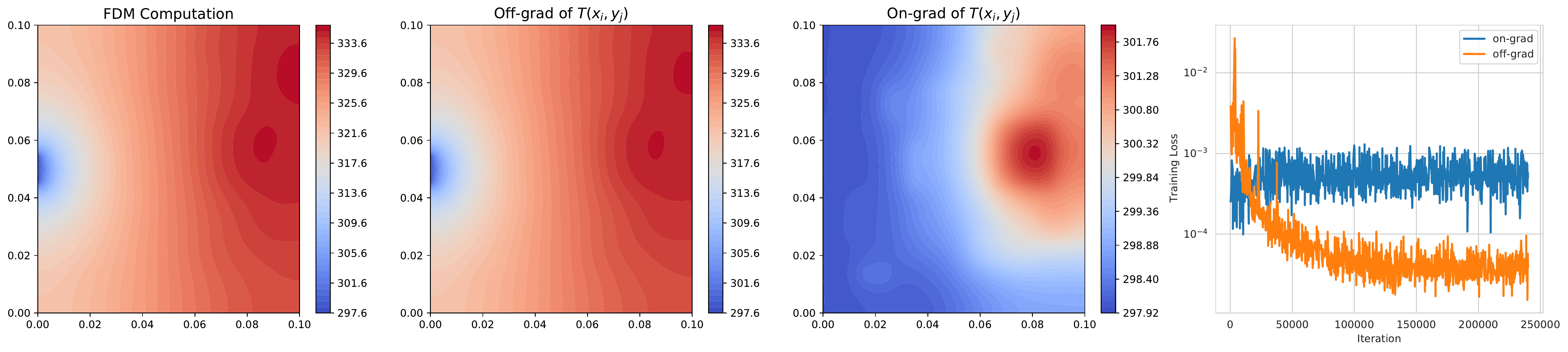}
	\caption{The effect of grad computation on or off for $T(x_i,y_j)'$.}
	\label{fig:on_off_grad}
\end{figure*}

\noindent \textbf{On-off of grad computation for $T(x_i,y_j)'$.} In the physics-informed loss Eq.~\ref{eq:12}, there is a intermediate variable $T(x_i,y_j)'$. The on-off of grad computation for $T(x_i,y_j)'$ has dramatically influenced the network optimization. 
The effect is shown in Fig.~\ref{fig:on_off_grad}. When turning on the grad computation for $T(x_i,y_j)'$, the network is hard to optimize, and it tends to output zero values (the prediction values are started at 298K).
The intermediate variable $T(x_i,y_j)'$ is constructed from surrounding predictions. When turning on the grad computation, the loss values in the current position will affect the adjacent position through back propagation, which may cause an abnormal trend of training loss.

\begin{figure*}[tp]
	\centering
	\subfigure[case 1]{
		\includegraphics[width=0.35\linewidth]{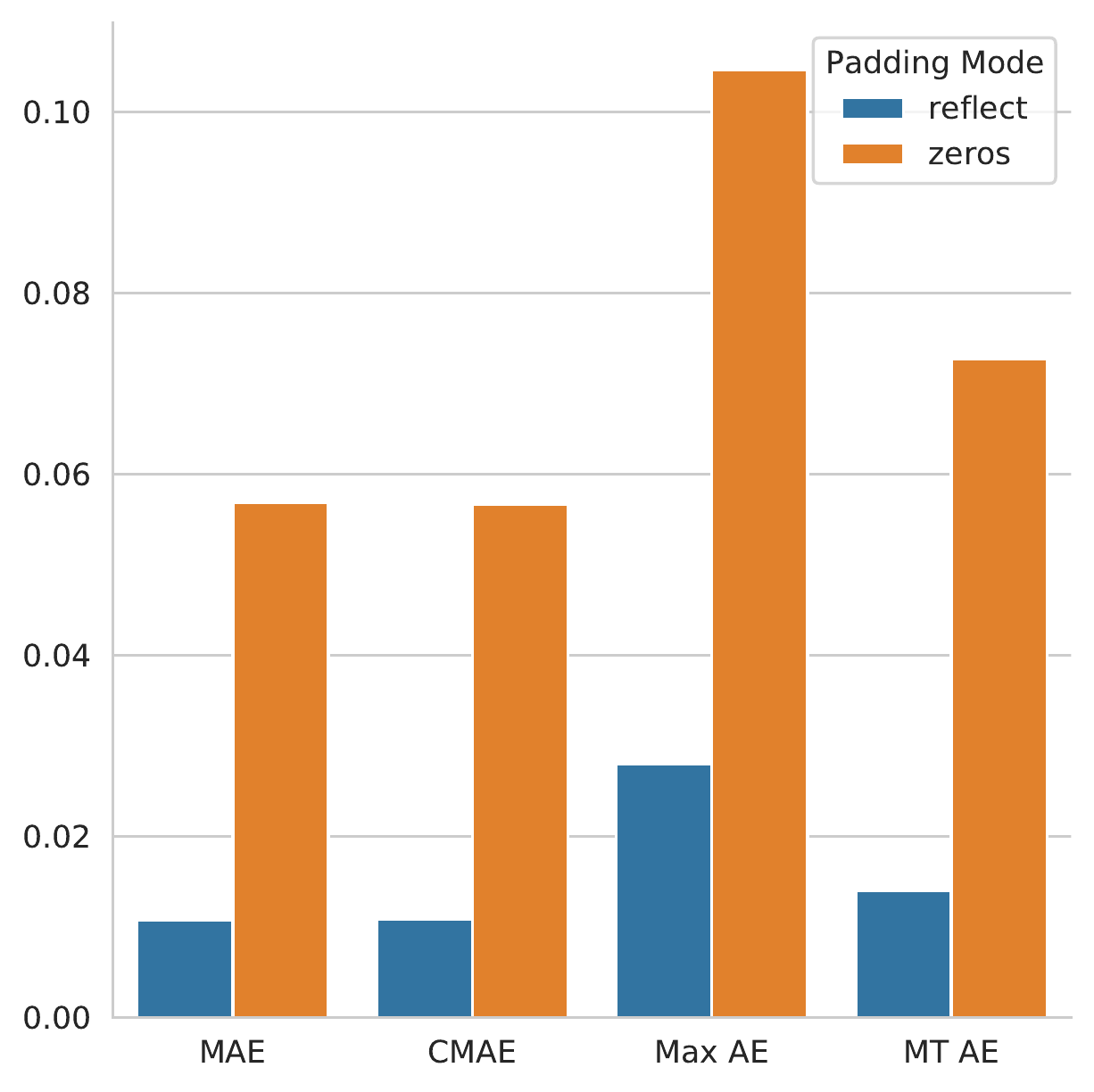}
	}
	\subfigure[case 2]{
		\includegraphics[width=0.35\linewidth]{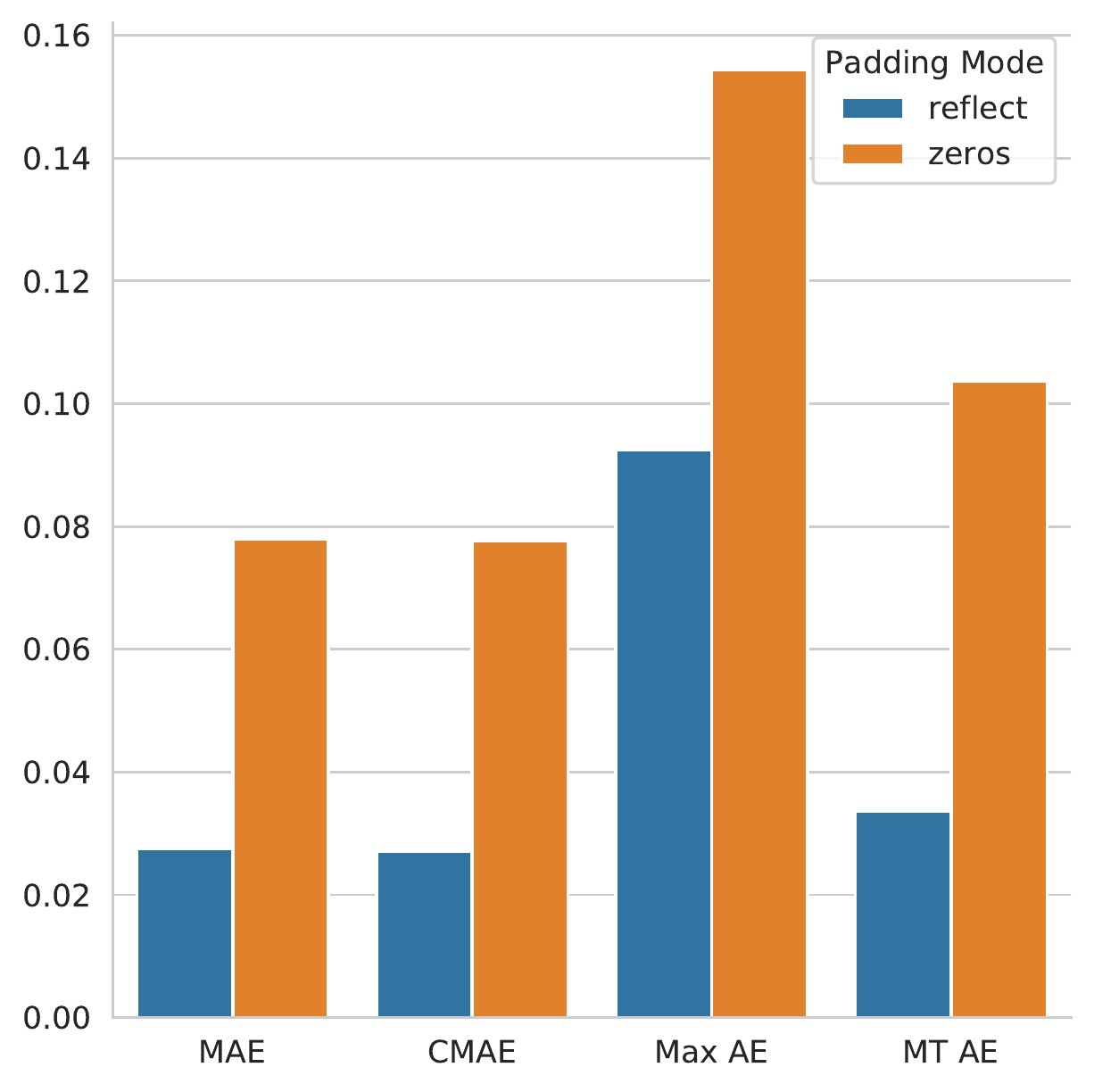}
	}
	\caption{The comparison of 'reflect' and 'zeros' padding mode.}
	\label{fig:padding_mode}
\end{figure*}

\begin{figure*}
	\centering
	\includegraphics[width=0.95\linewidth]{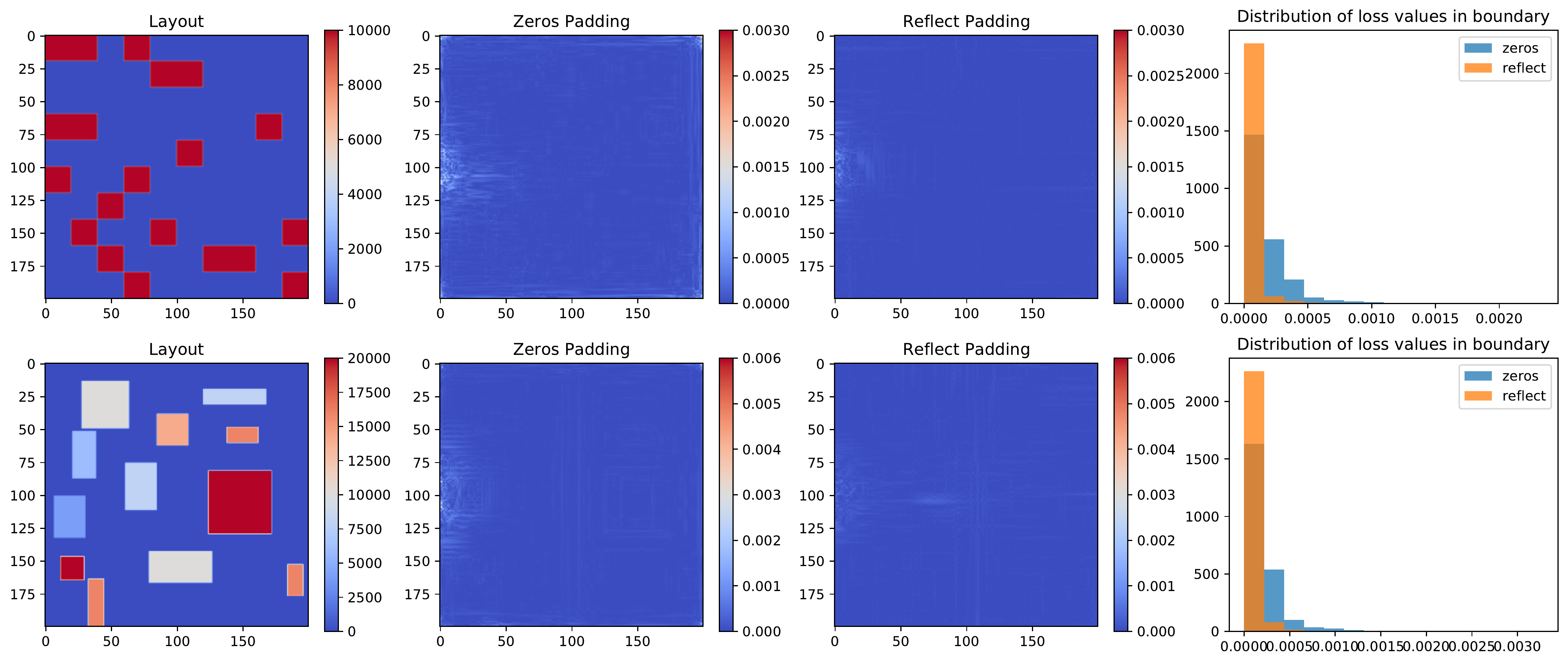}
	\caption{The distribution of physics-informed loss in 'reflect' and 'zeros' mode.}
	\label{fig:padding_dis}
\end{figure*}

\noindent \textbf{Padding mode.} In the convolution layer with 3$\times$3 filter, the padding operator is adopted to make the input and output size unchanged. As shown in Fig.~\ref{fig:padding_mode}, 'reflect' padding mode performs much better than 'zeros' in all evaluation metrics. 
The distribution of loss values in two padding modes can be seen in Fig.~\ref{fig:padding_dis}. 
The loss values are relatively larger in the boundary when using 'zeros' padding mode.
Since the physics-informed loss is based on finite difference, the prediction error will diffuse from the boundary, and the prediction precision in the whole domain is reduced.

\begin{figure*}[tp]
	\centering
	\subfigure[case 1]{
		\includegraphics[width=0.35\linewidth]{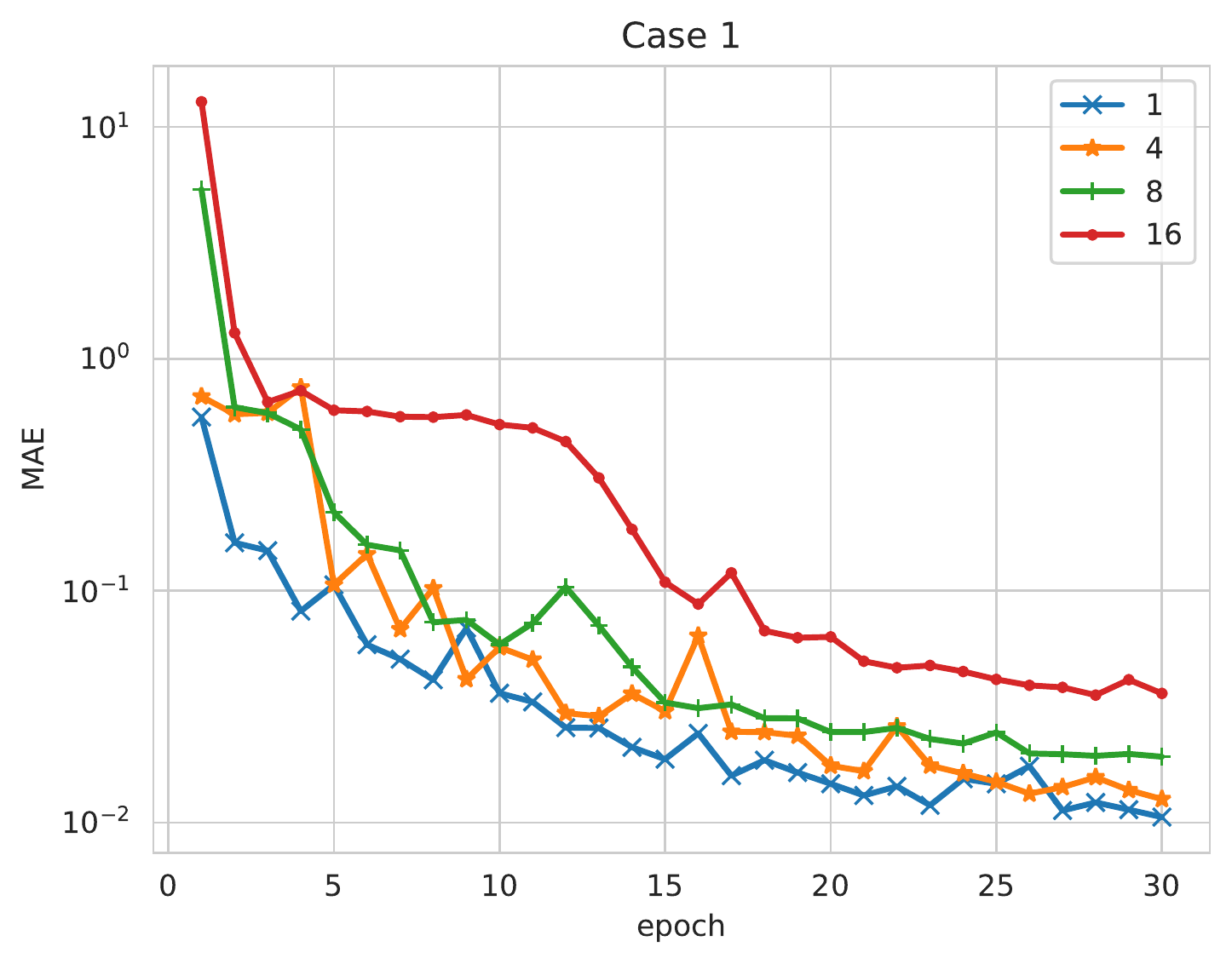}
	}
	\subfigure[case 2]{
		\includegraphics[width=0.35\linewidth]{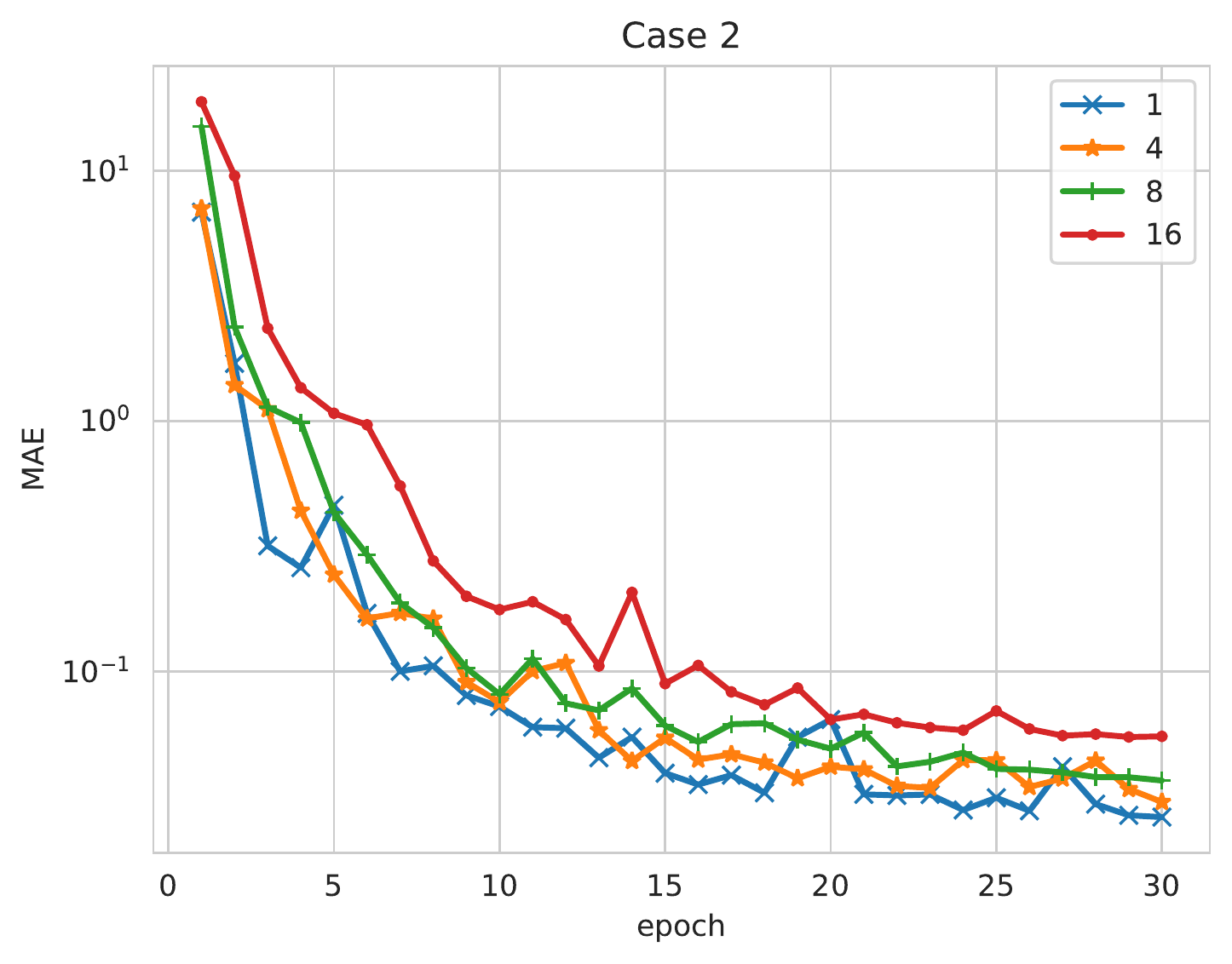}
	}
	\caption{The trend of validation MAE trained with different batch sizes. MAE is computed between the proposed method and FDM.}
	\label{fig:trend_bt}
\end{figure*}

\noindent \textbf{Batch Size.} Fig.~\ref{fig:trend_bt} shows the ablation study on the batch size. In computer vision tasks, a larger batch size usually benefits the network training. However, the results are better when the batch size is set to smaller for the task at hand.
In fixed training epochs, a smaller batch size means more iteration steps, contributing to the network convergence for HSL-TFP.

\begin{table}[htbp]
	\centering
	\caption{Comparison of various network components.}
	\begin{tabular}{ccccccc}
		\hline\noalign{\smallskip}
		& Normalization & Upsampling & Activation Function & Batch Size & Train MAE & Test MAE \\
		\noalign{\smallskip}\hline\noalign{\smallskip}
		\multirow{7}[0]{*}{Case 1} & GN    & Bilinear & GELU  & 1     & 0.0098 & 0.0108 \\
		& GN    & Bilinear & ReLU  & 1     & 0.0170 & 0.0191 \\
		& GN    & Bilinear & Tanh  & 1     & 0.0198 & 0.0214 \\
		& BN    & Bilinear & GELU  & 1     & 0.5623 & 0.5824 \\
		& BN    & Bilinear & GELU  & 16    & 0.4701 & 0.4943 \\
		& IN    & Bilinear & GELU  & 1     & 0.0184 & 0.0197 \\
		& GN    & Transpose & GELU  & 1     & 0.0409 & 0.0421 \\
		\noalign{\smallskip}\hline\noalign{\smallskip}
		\multirow{7}[0]{*}{Case 2} & GN    & Bilinear & GELU  & 1     & 0.0230 & 0.0275 \\
		& GN    & Bilinear & ReLU  & 1     & 0.0464 & 0.0515 \\
		& GN    & Bilinear & Tanh  & 1     & 0.0435 & 0.0491 \\
		& BN    & Bilinear & GELU  & 1     & 1.1383 & 1.1736 \\
		& BN    & Bilinear & GELU  & 16    & 0.8775 & 0.9358 \\
		& IN    & Bilinear & GELU  & 1     & 0.0340 & 0.0396 \\
		& GN    & Transpose & GELU  & 1     & 0.0964 & 0.0978 \\
		\noalign{\smallskip}\hline
	\end{tabular}%
	\label{tab:compar_com}%
\end{table}%

\noindent \textbf{Normalization methods, upsampling methods, and activation function.} We conduct ablation experiments to analyze the chooses of normalization, upsampling, and activation function.
As shown in Table~\ref{tab:compar_com}, the results of GN and IN are much better than BN, and GN is slightly better than IN. There is a great improvement of BN when turning on the training mode in evaluation. It means the statistics of BN are not incorrect in training. However, the results of BN are not notably improved with increasing the batch size to 16.
For two cases, bilinear upsampling yields significantly better performance than transpose convolution.
There is no clear difference in the performance of the three activation functions, but GELU is a little better.
The experiments demonstrate that the combination of GN, bilinear upsampling, and GELU is most effective.

\section{Conclusions}
\label{sec:5}
This paper investigates the physics-informed convolutional neural networks for temperature field prediction of heat source layout. 
First, we construct a physics-informed loss inspired by physics equations and the finite difference method. The loss guides the network to learn an operator mapping from intensity distribution function to solution function without simulation data.
Furthermore, we study to impose hard constrain on the Dirichlet and Neumann boundary to speed the network convergence.
Finally, we develop a well-designed architecture based on UNet for the HSL-TFP task, and pixel-level online hard example mining is proposed to balance different optimization difficulties in the computation domain.
For two HSL-TFP tasks, the experiments demonstrate that the proposed method can make high-precision temperature field predictions comparable to the numerical method and data-driven models.
The trained network can directly output the solution of HSL-TFP without solving PDE. It can be employed as a surrogate model to assist the layout optimization benefit from rapid and high-precision prediction. 

Similar to the steady-state temperature field prediction, the developed methods can be generalized to other physics problems, employing networks to solve parameterized PDEs, such as elastic mechanics, fluid mechanics, and electromagnetism. Furthermore, it is promising to investigate the methods in irregular computation domain and large-scale mesh. 
In some applications, data-driven method is more effective. Combining physics knowledge and labeled data is valuable to perform model verification, semi-supervised learning, and transfer learning.

\bibliography{bib.bib}
\end{document}